%% file: lrec-coling_latex.tex
\documentclass[10pt, a4paper]{article}

\usepackage{lrec-coling2024} 

\usepackage{times}
\usepackage{latexsym}
\usepackage[T1]{fontenc}

\usepackage[utf8]{inputenc}

\usepackage{microtype}
\usepackage{inconsolata}
\usepackage{multirow}
\usepackage{amsmath}
\usepackage{graphicx}
\usepackage{hyperref}

\title{\methodname: Narrative Reasoning by Grounding to Eventuality-centric Knowledge Graphs}

\newcommand{\ust}{\ensuremath{^\spadesuit}}
\newcommand{\amazon}{\ensuremath{^\diamondsuit}}

\name{Cheng Jiayang\ust, Lin Qiu\amazon, Chunkit Chan\ust, Xin Liu\ust, Yangqiu Song\ust, Zheng Zhang\amazon}

\address{\ust The Hong Kong University of Science and Technology, 
\amazon Amazon AWS AI\\
         \{jchengaj, yqsong\}@cse.ust.hk, 
         zhaz@amazon.com}

\abstract{
Narrative reasoning relies on the understanding of eventualities in story contexts, which requires a wealth of background world knowledge.
To help machines leverage such knowledge, existing solutions can be categorized into two groups.
Some focus on implicitly modeling eventuality knowledge by pretraining language models (LMs) with eventuality-aware objectives.
However, this approach breaks down knowledge structures and lacks interpretability.
Others explicitly collect world knowledge of eventualities into structured eventuality-centric knowledge graphs (KGs).
However, existing research on leveraging these knowledge sources for free-texts is limited.
In this work, we propose an initial comprehensive framework called \methodname, which aims to tackle the problem of grounding free-texts to eventuality-centric KGs for contextualized narrative reasoning.
We identify two critical problems in this direction: the \textit{event representation} and \textit{sparsity} problems.
We provide simple yet effective parsing and partial information extraction methods to tackle these problems.
Experimental results demonstrate that our approach consistently outperforms baseline models when combined with graph neural network (GNN) or large language model (LLM) based graph reasoning models.
Our framework, incorporating grounded knowledge, achieves state-of-the-art performance while providing interpretable evidence.
 \\ \newline  }

\begin{document}
\newcommand{\yq}[1]{\textcolor{red}{#1}}
\newcommand{\jy}[1]{{\color{blue} #1}}
\newcommand{\remove}[1]{{\color{red} #1}}

\newcommand{\draft}[1]{{\color{blue} #1}}
\newcommand{\methodname}{EventGround}

\maketitleabstract

\input{sections/introduction}
\input{sections/related_work}
\input{sections/methodology}
\input{sections/experiment}

\section{Conclusion}
\vspace{-8pt}

We point out two critical problems on grounding free-texts to eventuality-centric KGs, namely the \textit{event representation} and \textit{event sparsity} problems.
We propose a simple while effective approach, \methodname, to address these problems and to leverage the retrieved graph knowledge for narrative reasoning.
Empirical results demonstrate its consistent performance improvement.
Further investigation reveals that the normalization and partial information extraction components  drastically improve the grounding performance by alleviating event sparsity.

\newpage
\input{sections/limitation}

\section*{Acknowledgements}
The authors of this paper were supported by the NSFC Fund (U20B2053) from the NSFC of China, the RIF (R6020-19 and R6021-20) and the GRF (16211520 and 16205322) from RGC of Hong Kong. We also thank the support from the UGC Research Matching Grants (RMGS20EG01-D, RMGS20CR11, RMGS20CR12, RMGS20EG19, RMGS20EG21, RMGS23CR05, RMGS23EG08).

\bibliography{anthology,custom}
\bibliographystyle{acl_natbib}

\clearpage
\input{sections/appendix}

\end{document}

%% file: sections/introduction.tex
\section{Introduction}

Narrative reasoning, such as predicting story endings and reasoning with scripts, is a fundamental task in natural language understanding \cite{mostafazadeh2016corpus, li2018constructing, mori2020finding}. 
Reasoning with narratives depends on the understanding of eventualities\footnote{We use the linguistic term ``eventuality,'' which includes events, states, and activities \cite{mourelatos1978events, bach1986algebra}. For simplicity, we use the terms ``event'' and ``eventuality'' interchangeably.}\footnote{Jiayang completed this work while interning at Amazon AWS AI Lab.}.
Consider the following story:
\begin{quote}

``Tom was tired and wanted to have fun. He bought a movie ticket for Harry Potter.''

\end{quote}
It can be broken down into multiple sub-sentences:
\begin{quote}

(\textbf{E1}) Tom was tired. (\textbf{E2}) Tom wanted to have fun. (\textbf{E3}) He bought a movie ticket for Harry Potter.

\end{quote}
where each of them can be regarded as an \textit{event} with a verb and one to several arguments.
These events, which are considered as \textit{basic semantic units} in various NLP research \cite{zhang2020aser, yu2020cocolm, zhong2022unsupervised, zhang2022aser}, convey the majority of the meaning within their respective contexts.

\begin{figure}[t]
\centering
\includegraphics[width=0.5\textwidth]{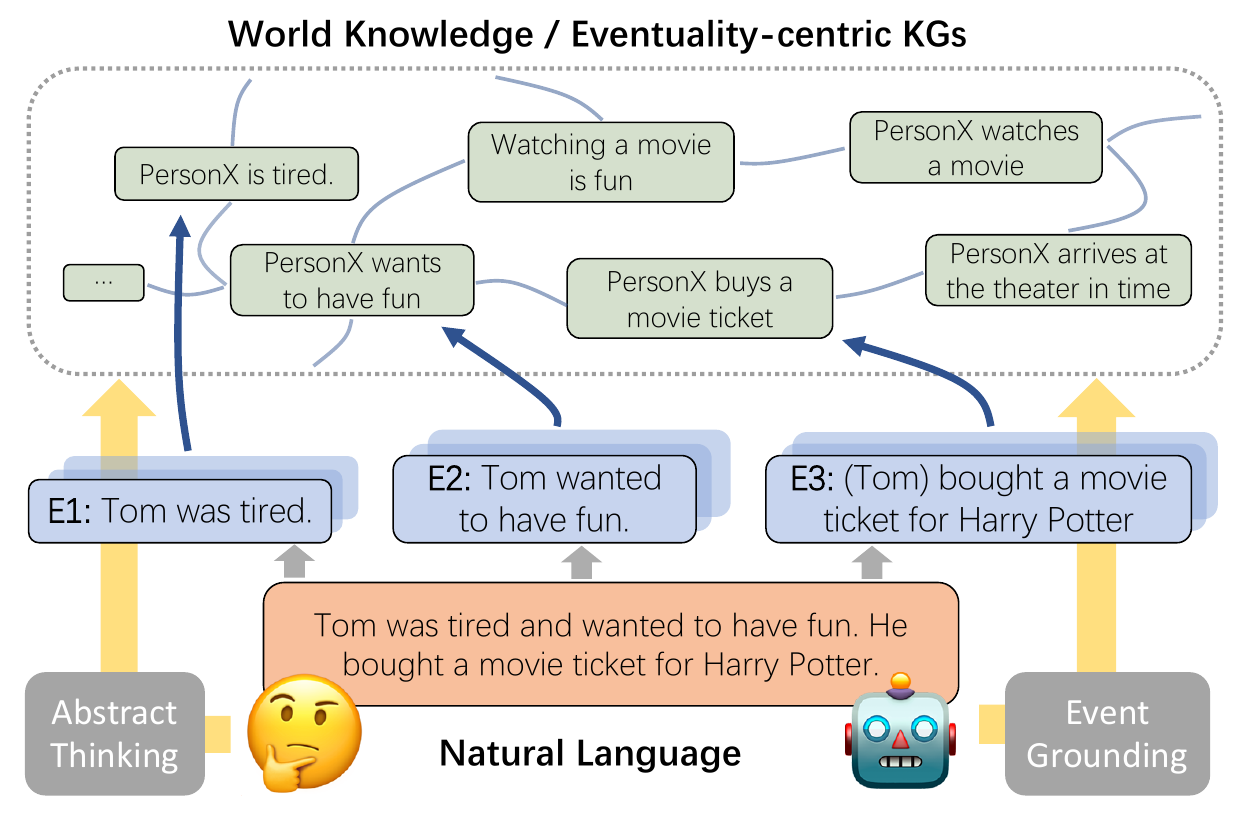}
\caption{Given a piece of story, our goal is to ground it to eventuality-centric KGs to retrieve contextualized background world knowledge for better narrative understanding.}
\label{fig:motivation}
\end{figure}

For human beings, the comprehension of these semantic units is found to heavily rely on our background \textit{world knowledge} beyond contexts \cite{day1998extensive}.
For instance, given \textbf{E1} and \textbf{E2}, we may infer that Tom might have just finished his work.
Since we know watching movies is a lot of fun, we find it reasonable that Tom chose to do so (from \textbf{E2} to \textbf{E3}).
We can also reason from \textbf{E3} that Tom would have to arrive at the theater before the movie started.


To model such world knowledge on machines, most existing work fall into two paradigms.
One is to implicitly model event knowledge by pretraining LMs with event-aware objectives \cite{yu2020cocolm, zhou2021modeling, zhou2022claret, zhou2022eventbert}.
This paradigm, however, sacrifices transparency and explanability of reasoning in its philosophy of design.
In comparison, another paradigm focuses on modeling the explicit symbolic event knowledge, usually in the form of eventuality-centric knowledge graphs (KGs, such as ASER \cite{zhang2022aser} and ATOMIC \cite{sap2019atomic}).
In this direction, how to leverage the symbolic event knowledge in these KGs for reasoning remains under-explored.
The handful research here only work on a restricted format (\textit{subject-verb-object}) of texts and could not generalize to free-texts \cite{li2018constructing, lv2020integrating, lee2019multi, lee2020weakly}.

In this paper, we make a step forward to examine the problem of grounding\footnote{Here, the term ``grounding'' refers to a process similar to ``linking'' used in ``entity linking'', where the target is the eventuality-centric KGs.} free-texts to eventuality-centric KGs.
This problem is non-trivial due to the distinct characteristics of events, including:

\begin{enumerate}
\vspace{-6pt}
    \item \textit{Difficulty in representing events.} 
    First, events appear entangled in texts.
    They tend to share arguments with other events in the same context (e.g., \textbf{E1} and \textbf{E2}).
    Second, when separated from the context, events lose co-reference information in the argument level.
    For instance, it is hard to discern whether the pronoun ``he'' in event \textbf{E3} refers to ``Tom'' in \textbf{E1} and \textbf{E2} or not.

\vspace{-6pt}
    \item \textit{Sparsity of events.} 
    Events are sparse in natural language.
    For instance, by adding or removing details, one could paraphrase \textbf{E3} into infinite events describing the same scenario, such as \textit{``he purchased a ticket online for the latest Harry Potter''} or \textit{``he booked a ticket''}.
    Given the incomplete nature of eventuality-centric KGs, matching arbitrary events to KGs has rather high failure rate.
\vspace{-6pt}
\end{enumerate}

To tackle the above problems, we propose the very first framework to explicitly ground free-texts to eventuality-centric KGs.
For the \textit{event representation} problem, we equip semantic parsing based event extraction with an event normalization module, which separates events from contexts while preserving co-reference information.
Motivated by humans' abstract thinking process, we propose a partial information extraction approach to tackle the \textit{sparsity} problem. This approach conceptualizes events into multiple partial events by omitting argument details.
Interestingly, we empirically demonstrate that these solutions significantly alleviate the sparsity problem.
Further, we ground the partial events to KGs to get joint reasoning subgraphs.
Subsequently, we employ two common graph reasoning models to leverage this knowledge. 
In addition to a model based on graph neural networks (GNN), we also utilize a model based on a large language model (LLM).
Experimental results on three narrative reasoning tasks show that our framework consistently outperforms current state-of-the-art models.
Lastly, we provide a qualitative study to showcase how our approach can provide interpretable evidence for model predictions.

To summarize, the paper's contributions are\footnote{The code and data are available at \url{https://github.com/HKUST-KnowComp/EventGround}.}: 
\begin{enumerate}
    \item We develop an initial formulation for the problem of grounding free-texts to eventuality-centric KGs. 
    \item We propose \methodname, a systematic approach, to solve the \textit{event representation} and \textit{sparsity} problems, and perform narrative reasoning based on the grounded information.
    \item Experimental results show that our approach outperforms strong baselines and achieves new state-of-the-art performance on three datasets, while providing human-interpretable evidence.
\end{enumerate}

%% file: sections/related_work.tex
\section{Related work}

Reasoning on narratives is a fundamental task~\cite{mostafazadeh2016corpus,li2018constructing,mori2020finding, jiayang2023storyanalogy} and has attracted much interest in the NLP community. 
The most crucial problem in narrative reasoning is modeling the relationship between events, which often requires background world knowledge~\cite{day1998extensive, mostafazadeh2016corpus}. 
Many large scale knowledge graphs (KGs) such as ATOMIC~\cite{sap2019atomic}, ConceptNet~\cite{speer2017conceptnet}, ASER~\cite{zhang2020aser,zhang2022aser} and GLUCOSE~\cite{mostafazadeh2020glucose} have been constructed in recent years.
Current solutions on leveraging the knowledge in these resources can be coarsely categorized into the following two groups.
An overview of the two paradigms is presented in Figure~\ref{fig:paradigm}.
The knowledge model paradigm leverages external KGs by pretraining LMs with carefully designed objectives.
Most existing knowledge enhanced LMs focused on using entity-centric KGs~\cite{DBLP:conf/acl/ZhangHLJSL19,DBLP:conf/emnlp/PetersNLSJSS19,DBLP:conf/emnlp/FevrySFCK20,DBLP:journals/corr/abs-2007-00849,DBLP:conf/iclr/XiongDWS20,DBLP:journals/corr/abs-1904-09223,DBLP:journals/corr/abs-2107-02137,DBLP:journals/tacl/JoshiCLWZL20}.
As for using external event knowledge, the knowledge model paradigm focus on finetuning language models on event-aware KGs, such as event-pair relation modeling \cite{DBLP:conf/acl/BosselutRSMCC19,DBLP:journals/corr/abs-2110-07178,zhou2021modeling}, whole event recovering/masking  \cite{zhou2022claret, yu2020cocolm}, and correlation-based event ranking \cite{zhou2022eventbert}.

The retrieval-and-integration paradigm, in contrast, explicitly retrieves triples or subgraphs from external KGs.
Recent work on reasoning with external KB and texts have explored grounding entities to KGs, such as \cite{sun2018open, sun2019pullnet, xiong2019improving, min2019knowledge, lee2021modeling}, and \cite{lin2019kagnet, feng2020scalable, yasunaga2021qa} in open-domain QA, commonsense QA, and narrative reasoning.
However, most of them ground to entity-centric KGs (e.g. the entity part of ConceptNet \cite{speer2017conceptnet}), which have little or no event knowledge.
Although some \cite{lv2020integrating, lee2019multi, lee2020weakly, li2018constructing} on script reasoning have investigated the usage of events, their methods are restricted to the ``subject-verb-object''-like structured texts in the MCNC task, and have difficulty extending to general free-texts.
In comparison, we tackle the more difficult problem of grounding events in free-texts to eventuality-centric KGs.
The wide adoption of AI critically needs explainability~\cite{hoffman2018metrics}. Thus, despite the appeal of a simpler pipeline (aided by the availability of large LMs), this work extends the retrieval-and-integration paradigm for grounding free-texts to eventuality-centric KGs for narrative reasoning.

 As opposed to event grounding, a similar term ``event linking'' has been used in the literature, where they either focus on cross-document event co-reference \cite{nothman2012event, krause2016event}, or event co-reference to Wikipedia pages \cite{yu2021event}. Moreover, their ``event'' refers to specific happenings such as ``World War II'' rather than the more general eventualities in this work.

%% file: sections/methodology.tex
\section{\methodname: Grounding free-texts to eventuality-centric knowledge graphs}
\label{sec:method}

\begin{figure}[t]
\centering
\includegraphics[width=0.5\textwidth]{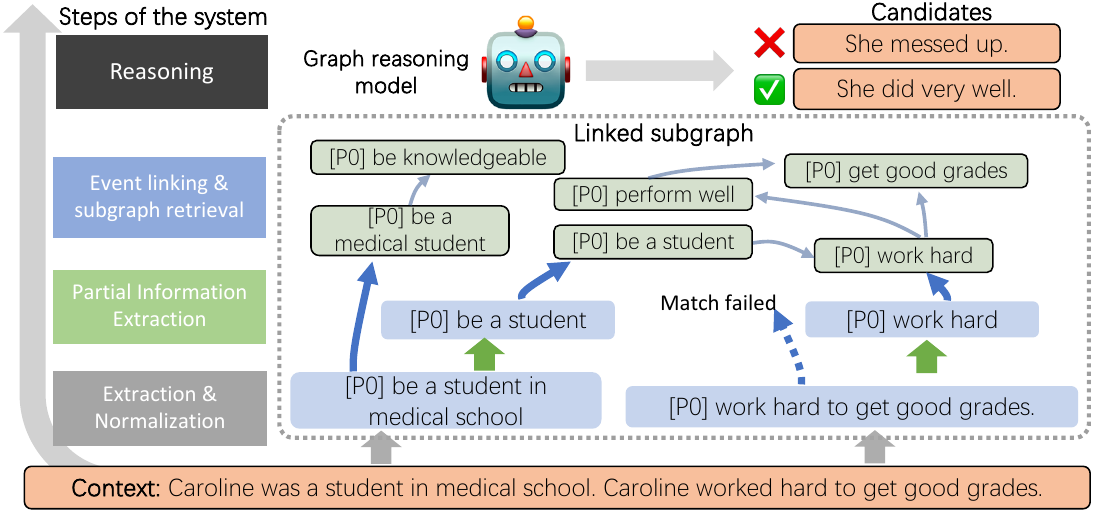}
\caption{An overview of \methodname.}
\label{fig:framework}
\end{figure}


In this section, we present our proposed framework, \methodname.
An overview is presented in Figure \ref{fig:framework}.
To tackle the \textit{event representation} problem, we equip semantic parsing based event extraction ($\S$~\ref{sec:event_extraction}) with an event normalization module ($\S$~\ref{sec:event_normalization}) to separate events from contexts while preserving their arguments' co-reference information.
We solve the \textit{sparsity} problem by with a partial information extraction approach ($\S$~\ref{sec:event_abstraction}).
We empirically prove that these solutions largely alleviate the sparsity problem in $\S$~\ref{sec:ablation}. 
At the end of this section, we discuss grounding the partial events to KGs to obtain joint reasoning subgraphs in $\S$~\ref{sec:event_grounding}, and present both the GNN-based and LLM-based reasoning models in $\S$~\ref{sec:graph_model}. 


\subsection{Obtaining events}
The proposed event acquisition pipeline includes event extraction ($\S$~\ref{sec:event_extraction}), normalization ($\S$~\ref{sec:event_normalization}) and partial information extraction ($\S$~\ref{sec:event_abstraction}).

\subsubsection{Event extraction}
\label{sec:event_extraction}
As shown in the previous example, events do not naturally exist in free texts.
Instead, an event may share arguments with (e.g., \textbf{E1} and \textbf{E2}) or contain another event.
Therefore, a special extraction step is needed to separate events from their contexts.

In this work, we consider the semantic parsing based methods to extract events from their contexts.
For each piece of text $s=[s_1, s_2, \cdots, s_n]$ with $n$ sentences, we conduct semantic role labeling (SRL) on the text to extract a series of verb-centric events $\mathcal{P}=\{p_1, p_2, \cdots, p_m\}$, where each event $p_i=(verb^i, \mathcal{A}^i)$ has a trigger $verb^i$ and a set of arguments $\mathcal{A}^i$.
Each argument $a_j^i\in \mathcal{A}^i$ has  a semantic role $role(a_j^i)\in \{ARG_0, ARG_1, \cdots, ARG_M\}$\footnote{The annotation follows the PropBank \cite{palmer2005proposition} annotation guideline, where the numbered arguments in general correspond to the roles: $ARG_0$-agent; $ARG_1$-patient; $ARG_2$-instrument, benefactive, attribute; $ARG_3$-starting point, benefactive, attribute; $ARG_4$-ending point; $ARG_M$-modifier.}.
In addition, we define the operator $text(p_i)$ to obtain the text of $p_i$.

\subsubsection{Event normalization}
\label{sec:event_normalization}
It is noteworthy that the extracted events suffer from the loss of co-reference information.
For instance, here are three events extracted from a text:\footnote{For simplicity, we do not explicitly show verbs and arguments of the events. All the words in events are lemmatized in our pipeline, which is not shown in the examples.}
\begin{quote}
    (1) The general had some wine at a party. \\
    (2) He felt sleepy. \\
    (3) He said goodbye to them.
\end{quote}
where ``\textit{the general}'' and ``\textit{he}'' refer to the same person, while ``\textit{them}'' refers to another group of people.
A system would not be aware of this co-reference relationship without contexts.
This makes it difficult to reason on the extracted events.

Motivated by previous work \cite{sap2019atomic, fang2021discos} in constructing commonsense KGs, we replace tokens referring to people with special tokens\footnote{Specifically, the spans of personal words are detected by syntactic parsing and animacy classification. We then employ the co-reference information between these spans to normalize all spans that refer to persons.} (e.g., ``\texttt{[P0]},'' ``\texttt{[P0's]},'' ``\texttt{[P1]},'' where different numbers refer to different people).
For instance, ``\textit{the general}'' and ``\textit{he}'' are replaced by ``\texttt{[P0]},'' and ``\textit{them}'' is replaced by ``\texttt{[P1]}.''
Through this normalization process, the co-reference information is preserved:
\begin{quote}
    (1) \texttt{[P0]} had some wine at a party. \\
    (2) \texttt{[P0]} felt sleepy. \\
    (3) \texttt{[P0]} said goodbye to \texttt{[P1]}.
\end{quote}

In addition, the normalization helps reduce event sparsity by removing details in the personal words.
For instance, ``\textit{the general felt sleepy},'' ``\textit{Joe felt sleepy},'' and ``\textit{he felt sleepy}'' will all be normalized to ``\textit{\texttt{[P0]} felt sleepy}.''
This increases their probability of being successfully grounded to KGs.

\subsubsection{Partial information extraction}
\label{sec:event_abstraction}

The normalized events retain rich contextual details from the original texts, which are important for downstream reasoning processes.
However, the sparsity of events can pose challenges in event grounding, especially when most knowledge graphs (KGs) are far from complete \cite{min2013distant, xiong2019improving}. 
For example, a KG is more likely to include a general event like ``\textit{a person is drinking}'' than ``\textit{the general is drinking Sauvignon Blanc on the balcony},'' because the former is more general and likely to occur frequently.

Humans strongly depend on conceptual abstraction to identify similarities among seemingly different concepts and events, which enables generalizations to unfamiliar situations \cite{murphy2004big}.
For instance, we can learn that there is common abstraction between ``\textit{buy a ticket for `Avengers'}'' and ``\textit{buy a ticket for `Harry Potter'},'' and that how the commonality ``\textit{buy a ticket}'' relates to other events such as we should ``\textit{arrive at the theater in time}''.
With this concept in mind, we use a partial information extraction (PIE) phase to obtain partial events as a method of controllable abstraction.

The partial information extraction is based on the importance of event arguments in semantic role labeling~\cite{palmer2005proposition}.
For instance, $ARG_0$ and $ARG_1$ have the highest importance as they usually specify the subject and objects.
In contrast, the modifier argument $ARG_M$ express the least information, as it usually defines additional constraints of the predicate, such as when and where the event happens.
Specifically, we propose to drop the event arguments in the descending order of their importance.
For event $p=(verb, \mathcal{A})$ with $|\mathcal{A}|=k$, we iteratively drop its argument $a_j\in\mathcal{A}$, such that the roles of dropped arguments follow the order: (1) $ARG_M$\footnote{We do not drop the negation (e.g., \textit{not}, \textit{n't}, \textit{never}) and modals (e.g., \textit{will}, \textit{may}, \textit{can}) modifier arguments, since they are crucial building blocks in discourse as revealed in the linguistics study \cite{jordan1998power}.}, (2) $ARG_2$, $ARG_3$, $ARG_4$, (3) $ARG_1$ and (4) $ARG_0$.
The partial information extraction on event set $\mathcal{P}$ results in a new set of partial events $\mathcal{P}_{abs}$, where $\mathcal{P}_{abs}=\{\hat p_1, \hat p_2, \cdots, \hat p_m\}$.
Each element $\hat p_i=[p_i^0, p_i^1, \cdots]$ is a sequence of partial events correspond to event $p_i\in\mathcal{P}$ ($p_i^0=p_i$).

Below is an example of $\hat p$:
\begin{enumerate}
   \item[$p^0$] ARG0: \underline{\texttt{[P0]}} V: \underline{evacuated} ARG2: \underline{to a relative 's house} ARGM: \underline{last night}. 
    \item[$p^1$] ARG0: \underline{\texttt{[P0]}} V: \underline{evacuated} ARG2: \underline{to a relative 's house}. 
    \item[$p^2$] ARG0: \underline{\texttt{[P0]}} V: \underline{evacuated}.
    \item[$p^3$] V: \underline{evacuated}. 
\end{enumerate}

Each time an argument is dropped, the abstract level of the partial event increases.
Meanwhile, partial events on higher abstract level (e.g. $p^2$, $p^3$) are more likely to have been recorded in KGs, which alleviates the sparsity problem.
In $\S$~\ref{sec:ablation}, we empirically show that the partial information extraction improves the model performance by drastically increasing the hit rate of event grounding.

\subsection{Grounding to eventuality-centric KG} 
\label{sec:event_grounding}

In this section, we discuss the event grounding approach.
In $\S$~\ref{sec:event_matching}, we describe how to map events to eventuality-centric KGs to get the anchor events that have the closest semantic meaning.
In $\S$~\ref{sec:subgraph}, we describe how to retrieve grounded subgraphs based on the anchor events.

\subsubsection{Event matching}
\label{sec:event_matching}

Suppose we have an eventuality-centric KG $\mathcal{G}=(\mathcal{V}, \mathcal{E})$.
$\mathcal{V}$ and $\mathcal{E}$ are the node set and the edge set, respectively.
Each node $v_i \in \mathcal{V}$ is an event with a text attribute $text(v_i)$.
Then, for each event $p\in\mathcal{P}_{abs}$, our goal is to find the node $v\in\mathcal{V}$ (which we term as ``\textit{anchor event}'') that is the most similar to $p$:
\begin{equation}
    v = \arg \min\limits_{v\in\mathcal{V}} d(p, v),
\end{equation}
where $d(\cdot,\space\cdot)$ denotes the distance between events.

To define the similarity, previous work have explored \textit{token-level similarity} by computing the cosine distance for TF-IDF or BM25 vectors \cite{lv2020integrating}.
However, this method overlooks the semantics of events, and constantly fails by mapping to events with high inverse document frequency terms (e.g. ``\textit{\texttt{[P0's]} \underline{lung} gets punched}'' is matched with ``\textit{\texttt{[P0]} has \underline{lung} cancer}'').
Therefore, we turn to use \textit{semantic similarity} to match events.

Specifically, we encode event $p$ and $v$ with sentence transformers \cite{reimers2019sentence},\footnote{\url{https://huggingface.co/sentence-transformers/all-MiniLM-L6-v2}} and compute $d(p,\space v)$ by the L2 distance:

\begin{equation}
    d(p, v) = ||\textrm{SBERT}(text(p)), \textrm{SBERT}(text(v))||_2.
\end{equation}

In practice, not every event can be successfully matched with the correct ones.
We empirically set a threshold $l$ over $d(p, v)$ to filter out the failed matches.\footnote{We sample 100 matching results and empirically set $l$=0.65 that filters out the most failed cases.}
As a result, partial events in $\mathcal{P}_{abs}$ are matched to their anchor events in $\mathcal{G}$, which we denote by $\mathcal{C}$.
$\mathcal{C}=\{\hat c_1, \hat c_2, \cdots, \hat c_m\}$, where each $\hat c_i$ is a sequence of anchor events matched from $\hat p_i$.

\subsubsection{Joint subgraph construction}
\label{sec:subgraph}

\noindent \textbf{Knowledge subgraph retrieval}
Based on the anchor events from the matching results in $\S$~\ref{sec:event_matching}, we aim to retrieve a subgraph $\mathcal{G}_{sub}=(\mathcal{V}_{sub}, \mathcal{E}_{sub})$ from $\mathcal{G}$.
Ideally, $\mathcal{G}_{sub}$ should contain the background world knowledge related to the reasoning, meanwhile cover minimal number of additional eventualities.
Finding such a subgraph is essentially trying to solve an NP-complete Steiner tree problem \cite{garey1977rectilinear, lin2019kagnet}, which is intractable.
As a workaround, we search for the shortest path within $\gamma$-hops between each event pair in $\{(v_a, v_b): v_a\in \hat c_i, v_b\in \hat c_j; \hat c_i, \hat c_j\in \mathcal{C}\}$.
For any path obtained, the nodes and edges along the path are added to $\mathcal{G}_{sub}$.

\noindent \textbf{Joint subgraph construction}
Based on $\mathcal{G}_{sub}$, we construct a joint knowledge enhanced subgraph $\mathcal{G}_{joint}=(\mathcal{V}_{joint}, \mathcal{E}_{joint})$ for reasoning.
Specifically, $\mathcal{G}_{joint}$ includes all the nodes and edges in $\mathcal{G}_{sub}$.
In addition, we add the context events in $\mathcal{P}$ as nodes to $\mathcal{G}_{joint}$, where their grounding relation to anchor events in $\mathcal{C}$ as well as the context relation (between the previous and latter events in the order that they appear in context) are added as edges.

\subsection{Graph reasoning models}
\label{sec:graph_model}

The retrieved subgraphs are then used for reasoning using either a GNN-based reasoning model or an LLM-based reasoning model.

\noindent \textbf{GNN-based reasoning model.}
We first encode the text $s$ and node $v\in\mathcal{V}_{joint}$ using the language model representation:
\begin{equation}
\begin{split}
    \textbf{v} & = f_{\small{\textsc{LM}}}(text(v)), \\
    \textbf{s} & = f_{\small{\textsc{LM}}}(s).
\end{split}
\end{equation}
Then, we employ a GNN module to perform reasoning on the joint subgraph $\mathcal{G}_{joint}$.
We choose the relational graph convolutional networks (RGCN) \cite{schlichtkrull2018modeling} so that the relational information in $\mathcal{G}_{joint}$ can be well modeled.
Specifically, for each layer $l$ in an $L$-layer GNN, the representation $\textbf{h}_i^{(l)}$ of node $i\in\mathcal{V}_{joint}$ is updated by
\begin{equation}
    \mathbf{h}_{i}^{(l+1)} = \sigma \Big(\sum\limits_{r\in\mathcal{R}}\sum\limits_{j\in\mathcal{N}_r(i)}\frac{1}{|\mathcal{N}_r(i)|}\mathbf{W}_r\cdot \mathbf{h}_{j}^{(l)}\Big),
\end{equation}
where $\mathcal{R}$ is the set of edge types in $\mathcal{E}_{joint}$, $\mathcal{N}_r(i)$ denotes the neighborhood with relation $r$ of node $i$, and $\sigma (\cdot)$ is an non-linear activation.
Then, we obtain the vector representation for $\mathcal{G}_{joint}$ by pooling the hidden node embeddings from the last layer
\begin{equation}
    \mathbf{g} = \textrm{Pooling}(\{\mathbf{h}_{i}^{L}: i\in\mathcal{V}_{joint}\}).
\end{equation}
The final prediction comes from 
\begin{equation}
    p(s) \propto \textbf{MLP} (\mathbf{s}+\mathbf{g}),
\end{equation}
where $\textbf{MLP}$ means a multi-layer perceptron module to predict the probability of the output.

\vspace{0.05\linewidth}

\noindent \textbf{LLM-based reasoning model.}
Recently, LLMs have been employed to reason over symbolic graphs \cite{xiong2024large, yang2023harnessing} to perform narrative reasoning.
In light of this, we also explored fusing the eventuality knowledge subgraph $\mathcal{G}_{joint}$ into LLMs.
Since LLMs only receive sequence inputs, we conduct sequentialization on subgraphs in a format similar to \cite{madaan-yang-2021-neural, sakaguchi-etal-2021-proscript-partially}.
Using a transformation function $t(\cdot)$, a subgraph $\mathcal{G}_{joint}$ is transformed into a piece of text $s_{\mathcal{G}_{joint}}$ ($s_{\mathcal{G}_{joint}}=t(\mathcal{G}_{joint})$), which is then fed into LLM as part of the prompts.
We discuss variations of $t(\cdot)$ and other details in $\S$~\ref{sec:exp-setup}.

%% file: sections/experiment.tex
\section{Experiments}

\subsection{Datasets}

We conduct experiments on three downstream tasks on narrative reasoning.
The statistics are presented in Table \ref{tab:dataset_statistics}.

\noindent$\bullet$ \textbf{Story Cloze Test v1.0} (SCT-v1.0) was proposed by \citet{mostafazadeh2016corpus} to evaluate the understanding of relations between events. 
Given four consecutive sentences, the task is to predict the correct ending from two possible choices.

\noindent$\bullet$ \textbf{Story Cloze Test v1.5} (SCT-v1.5) Later, \citet{sharma2018tackling} introduces a new version to correct the artifacts in the previous release.
For both versions, we follow the common practice \cite{li2019story, yu2020cocolm} to randomly select $100$ validation samples for validation, and use the rest for training.
    
\noindent$\bullet$ \textbf{Multiple Choice Narrative Chain} (MCNC) \cite{granroth2016happens, li2018constructing} is a 5-way multiple choice task that requires a system to predict the ending event given its previous context event sequence.

\begin{table}[t]
\centering
{\small
\begin{tabular}{lccc}
\hline
\textbf{Name} & \textbf{Train} &  \textbf{Valid} & \textbf{Test}\\
\hline
SCT-v1.0 & 1,771 & 100 & 1,871 \\
SCT-v1.5 & 1,471 & 100 & 1,571 \\
MCNC & 140,331 & 10,000 & 10,000 \\
\hline
\end{tabular}
}
\caption{Statistics of datasets.}
\label{tab:dataset_statistics}
\end{table}

\subsection{Eventuality-centric knowledge graphs}
 
There are eventuality-centric KGs such as ATOMIC \cite{sap2019atomic}, GLUCOSE \cite{mostafazadeh2020glucose} and ASER \cite{zhang2020aser, zhang2022aser}.
In this paper, we conduct experiments on ASER.
The nodes in ASER are eventualities, and the edges between them are the discourse relations (e.g.  ``Precedence'', ``Contrast'' and ``Reason'') defined in Penn Discourse Tree Bank \cite{prasad2008penn}.
To enable grounding normalized events to KGs, we normalize and aggregate eventualities in the ASER-core-100 version\footnote{We obtain the core-100 version by filtering out nodes with frequency lower than 100 from ASER-core: \url{https://hkust-knowcomp.github.io/ASER/}} by detecting and replacing the personal words with aforementioned special tokens.
The resulting normalized ASER graph contains $193k$ nodes and $6.6m$ edges.

\begin{table*}[t]
\centering
{\small
{
\begin{tabular}{ll|ccc}
\hline
\textbf{Method} & \textbf{Size} &  \textbf{SCT-v1.0} & \textbf{SCT-v1.5} & \textbf{MCNC}\\
\hline
\cite{lv2020integrating} & 125M & - & - & 58.66  \\
\cite{zhou2021modeling} & 469M & - & - & 63.62 \\
CoCoLM \cite{yu2020cocolm} & 355M & \underline{97.70} & - & - \\
TransBERT \cite{li2019story} & 355M & 91.80 & 90.30 & - \\
EventBERT \cite{zhou2022eventbert} & 355M & - & \underline{91.33} & 63.50 \\
ClarET \cite{zhou2022claret} & 400M & - & 91.18 & \underline{64.61} \\
\hline
RoBERTa-base \cite{liu2019roberta} & 125M & 92.75\small{$\pm$0.24} & 87.14\small{$\pm$0.39} & 61.28\small{$\pm$0.14}  \\
RoBERTa-large \cite{liu2019roberta} & 355M & 96.74\small{$\pm$0.08} & 92.34\small{$\pm$0.06} & 63.01\small{$\pm$0.12}  \\
DeBERTa-large \cite{he2021debertav3} & 354M & 98.13\small{$\pm$0.34} & 94.67\small{$\pm$0.25}  & 65.67\small{$\pm$0.13}  \\
\hline
\methodname\small{-RoBERTa-base} & 126M &  93.30\small{$\pm$0.11} & 87.65\small{$\pm$0.13} & 62.11\small{$\pm$0.07}   \\
\methodname\small{-RoBERTa-large} & 358M &   97.10\small{$\pm$0.13} & 92.86\small{$\pm$0.05} & 63.96\small{$\pm$0.15}   \\
\methodname\small{-DeBERTa-large} & 358M &   \textbf{98.29\small{$\pm$0.16}} & \textbf{95.01\small{$\pm$0.32}} & \textbf{66.05\small{$\pm$0.12}}  \\
\hline
\end{tabular}
}
}
\caption{Main results on the benchmarks. Numbers are mean and standard deviation of accuracy (\%) over three runs. \underline{Underlined results} are the previous state-of-the-art performance.}
\label{tab:main_result}
\end{table*}

\begin{table}[!t]
\small
\centering
\begin{tabular}{l c c}
\hline
\multicolumn{1}{c}{\textbf{Model}}
& \textbf{SCT-v1.0} & \textbf{SCT-v1.5}\\
\hline
Random                          & 50.00 & 50.00\\
ChatGPT$_\text{Vanilla}$         & 77.80 & 77.00 \\
ChatGPT$_\text{DOT}$ & 67.80 & 69.00 \\
ChatGPT$_\text{Node}$         & 72.00 & \textbf{78.00} \\
ChatGPT$_\text{Node \& Edge}$ & \textbf{79.60} & \textbf{78.00}\\
\hline
\end{tabular}
\caption{ChatGPT evaluation results (accuracy \%). We report the model performance when (1) ChatGPT$_\text{Vanilla}$: no knowledge is provided; (2) ChatGPT$_\text{DOT}$, ChatGPT$_\text{Node}$, and ChatGPT$_\text{Node \& Edge}$: the knowledge subgraphs are transformed into sequences as part of the inputs.}
\label{tab:ChatGPT_Performance}
\end{table}

\subsection{Experimental Setup}
\label{sec:exp-setup}
We implement the event extractor with AllenNLP SRL tools.\footnote{\url{https://github.com/allenai/allennlp}}
To normalize the events, the syntactic parser, animacy classifier, and co-reference tools are from Stanford CoreNLP.\footnote{\url{https://stanfordnlp.github.io/CoreNLP/}} 
In our implementation of the event matching module, due to the large scale of $|\mathcal{V}|$, we employ Faiss \cite{johnson2019billion} to accelerate the similarity search.
When retrieving subgraph, we set the shortest path length limit $\gamma$ to 3, meaning that there are at most 2 intermediate nodes between any two anchor nodes along the path.

We implement the GNN-based reasoning model with Deep Graph Library \cite{wang2019dgl} and Huggingface-Transformers \cite{wolf-etal-2020-transformers}.
For finetuning the supervised models, we conduct grid-search over model hyper-parameters. 
The number of convolutional layers $L$ are searched within $\{2, 3, 4\}$, and the hidden size of convolutional layers $\in\{64, 128, 256, 512\}$. 
For relational convolutional layers, the number of bases is searched within $\{-1, 10, 30\}$.
We use the Adam \cite{kingma2015adam} optimizer with cosine learning rate schedule to optimize the models.
The learning rate is set to $1e-5$ for all the ``base'' models, and $5e-6$ for all the ``large'' models.
All the experiments are run on 4 NVIDIA Tesla-V100 GPUs.

For the LLM-based reasoning model, we adopt ChatGPT~\cite{openai2022chatgpt}.~\footnote{The evaluation is performed in September 2023.}
We consider three implementations for the graph sequentialization function $t(\cdot)$: (1, DOT) using the DOT language to represent graphs \cite{gansner1993technique, madaan-yang-2021-neural, sakaguchi-etal-2021-proscript-partially}; (2, Node \& Edge) instead of using node indexing as in DOT, we try directly inputing all the nodes and edges (e.g., ``\texttt{[P0] buy a boat --> [P0's] nearby marina have a race; [P2] prepare --> [P2] go to sleep; ...}''); (3, Node) only the nodes are fed into ChatGPT (e.g., ``\texttt{[P0] buy a boat; [P0's] nearby marina have a race ...}'').
The prompt template is:
``\texttt{Event knowledge on narrative choice A: \{$t(\mathcal{G}_{joint, A})$\} \textbackslash n Event knowledge on narrative choice B: \{$t(\mathcal{G}_{joint, B})$\} \textbackslash n Question:\{\} \textbackslash n Answer:''}.
As a baseline, we also test ChatGPT without the additional knowledge (denoted by ``ChatGPT$_\text{Vanilla}$'').
For SCT-v1.0, we report results on its test set (sampled 500 instances). 
Since the test set of SCT-v1.5 is no longer publicly available\footnote{\url{https://competitions.codalab.org/competitions/15333}} at the time we ran this experiment, we report the results on its validation set.
We do not report the performance on MCNC because the lengths of most instances in this set exceed the maximum input length.

\subsection{Main results}

The main results on the three datasets are presented in Table \ref{tab:main_result} and \ref{tab:ChatGPT_Performance}. 
Per-task performance comparisons are presented in Appendix \ref{sec:appendix_detail_results}.

As shown in Table \ref{tab:main_result}, when coupled with a GNN-based reasoning model, our proposed framework achieves consistent performance gain over different backbone models.
Moreover, compared with existing knowledge enhanced models, we achieve SOTA performance in three narrative reasoning tasks.
The knowledge also benefits our LLM-based reasoning model (Table~\ref{tab:ChatGPT_Performance}), especially when the subgraphs are transformed using the ``$\text{Node \& Edge}$'' setting.

\subsection{Ablation study}
\label{sec:ablation}
We conduct ablation studies to investigate the contribution of each component in our framework.

\begin{table}[t]
\centering
{\small
\begin{tabular}{lcc}
\hline
     & \methodname\tiny{-RB} &  \methodname\tiny{-BB} \\
\hline
w/o know. & 92.75\small{$\pm$0.24} & 83.63\small{$\pm$1.16} \\
w/o extract. & 91.86\small{$\pm$0.21} & 83.74\small{$\pm$0.38} \\
w/o norm.   & 92.43\small{$\pm$0.46} & 83.98\small{$\pm$0.87} \\
\hline
w/o PIE & 92.81\small{$\pm$0.32} & 83.88\small{$\pm$1.40} \\
- ARGM & 93.17\small{$\pm$0.25} & 84.79\small{$\pm$1.37} \\
- ARG2,3,4 & 93.03\small{$\pm$0.49} & 84.53\small{$\pm$0.60} \\
- ARG1 & \textbf{93.30\small{$\pm$0.11}} & \textbf{85.78\small{$\pm$0.74}} \\
\hline
\end{tabular}
}
\caption{Effect of event extraction, normalization and partial information extraction (PIE). The mean and standard deviation of accuracies on SCT-v1.0 are reported, where ``RB'' and ``BB'' refer to RoBERTa-base and BERT-base versions.}
\label{tab:main_ablation}
\end{table}

\subsubsection{Effect of event extraction, normalization, and partial information extraction}
As shown in Table \ref{tab:main_ablation}, we ablate the event extraction (``w/o extract.''), the event normalization (``w/o norm.'') and the partial information extraction (``w/o PIE'' and ``- ARGX'') respectively.
Specifically, when ablating the event extraction module, we instead use the whole sentence for event grounding.
When ablating the event normalization part, we skip the normalization step, and use the raw events for grounding.
For partial information extraction, we drop event arguments in the order described in $\S$~\ref{sec:event_abstraction}, where the highest level (``- ARG1'') contains all the partial events in the previous levels.
The baseline (``w/o know.'') shows the results of vanilla language models, which do not leverage any external knowledge.

We have several observations. 
First, the event extraction and normalization steps are necessary. 
When removed, the performance relative to the baseline does not improve, or even drops.
Second, the partial information extraction step is crucial.
By only taking the first level of partial events (removing modifier arguments), we have seen considerable performance gain.
The model reaches its best performance after dropping ARG1.

In $\S$~\ref{sec:method}, we discuss the \textit{sparsity} of events.
Here, we conduct both automatic and human evaluation to discuss how our method contribute to the alleviation of sparsity.

\begin{figure}[ht]
\centering
\includegraphics[width=0.5\textwidth]{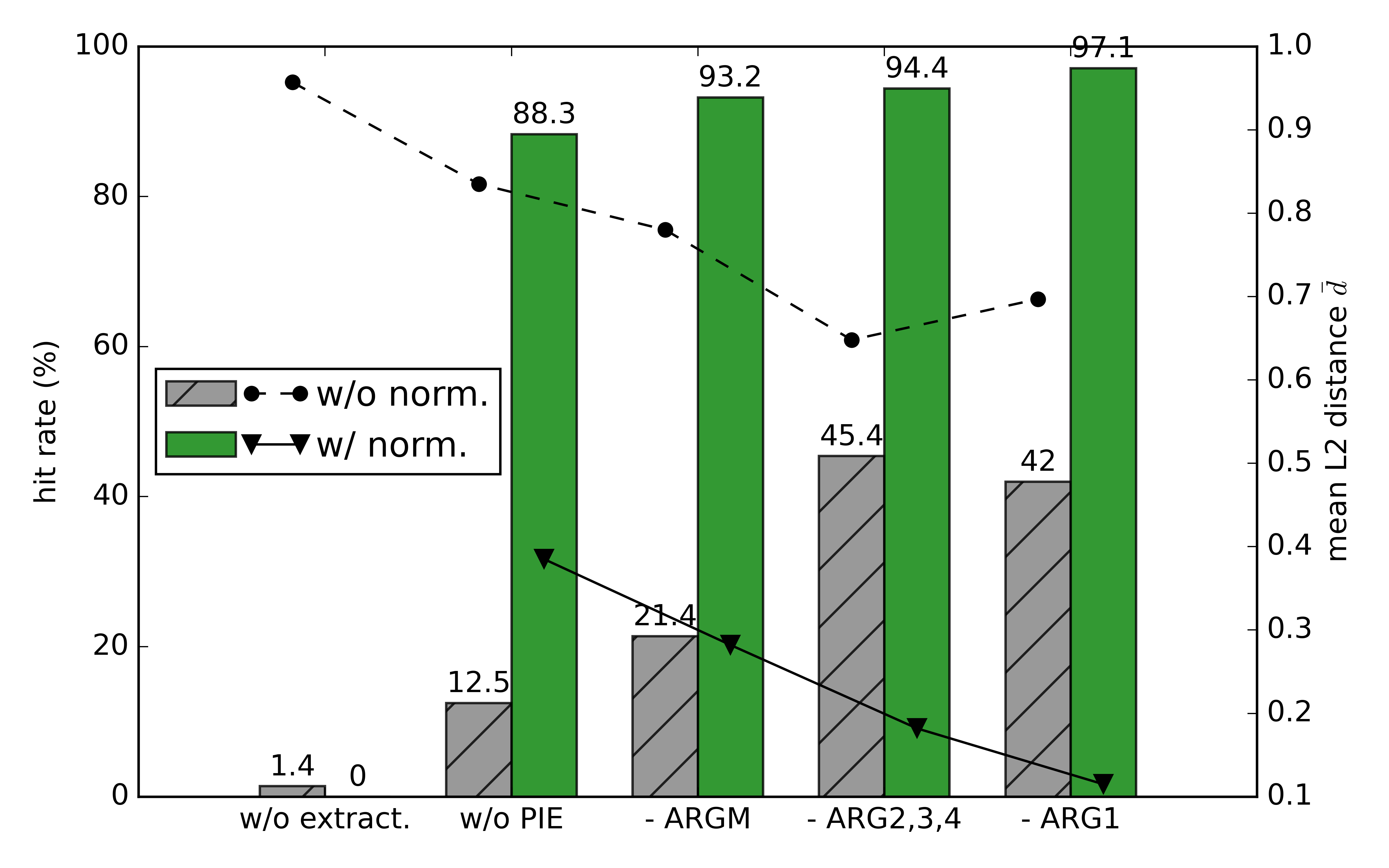}
\caption{A comparison on the event grounding performance under different settings. The bar plot (with $y$-axis on the left) shows the percentage hit rate of event matching. The lines show the average L2 distance $\bar d$. 
We do not conduct normalization for ``w/o extract.''. }
\label{fig:ablation_matching}
\end{figure}

\noindent$\bullet$ \textbf{Automatic Evaluation} (Figure \ref{fig:ablation_matching}) We analyze by automatic measures: (1) the average L2 distance $\bar d$ in event matching ($\S$~\ref{sec:event_matching}), and (2) the percentage of events considered as successful match, i.e. with L2 distance below $l=0.65$ (hit rate).

\begin{figure}[ht]
\centering
\includegraphics[width=0.5\textwidth]{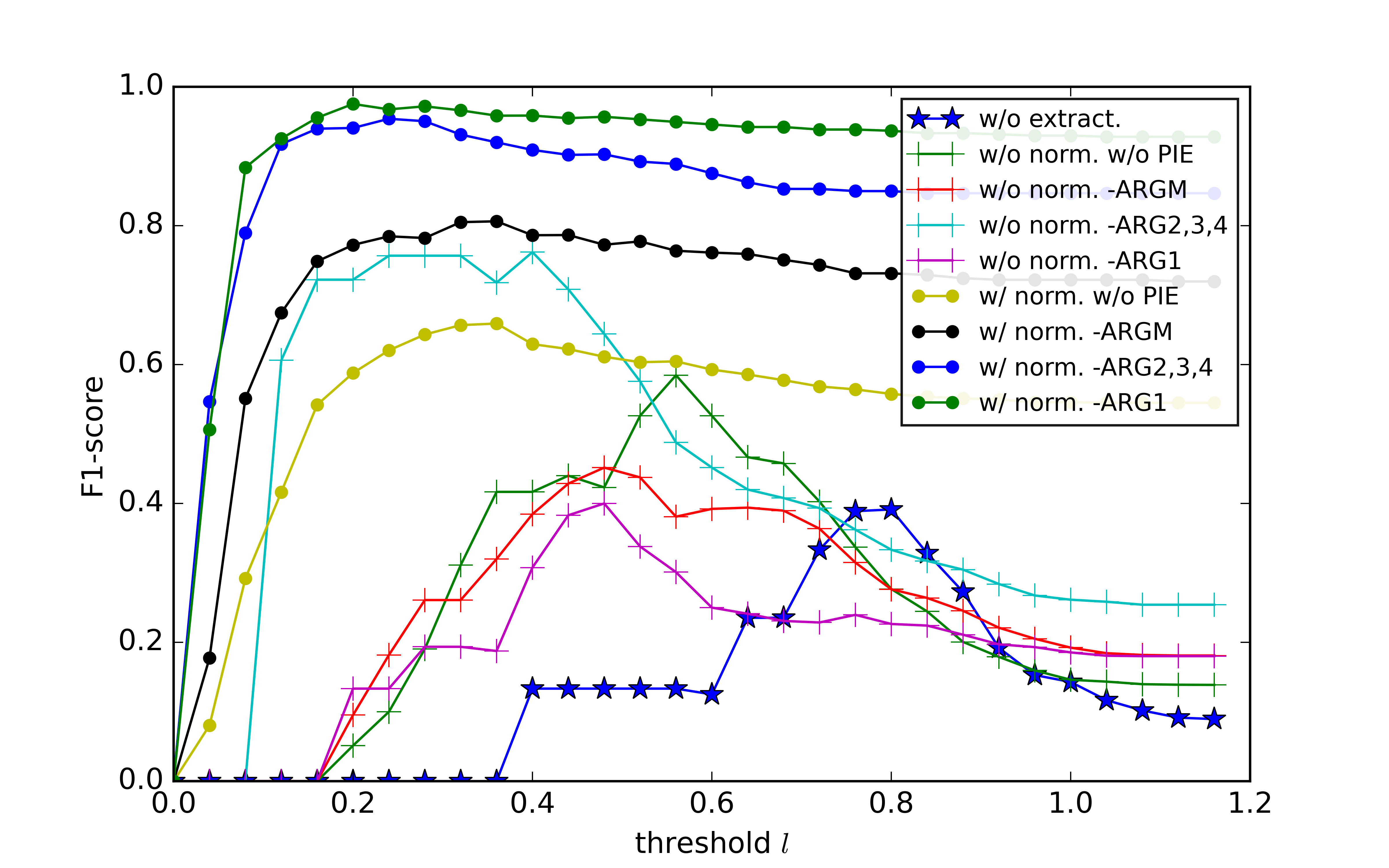}
\caption{The F1-score to threshold curves. They reflect the event matching performance under different threshold $l$. }
\label{fig:ablation_f1}
\end{figure}

\noindent$\bullet$ \textbf{Human Evaluation} (Table \ref{tab:human_eval}, Figure \ref{fig:ablation_f1}) 
    We evaluate the matching results by human annotation.
    Three domain experts are asked to annotate whether event matching is successful for 50 stories ($\sim$500 events) randomly sampled from the validation set of SCT v1.0. 
    The Fleiss's Kappa value is $0.7414$.
    We obtain ground-truth labels by majority vote, and present the accuracy of different event matching methods in Table \ref{tab:human_eval}.
    To investigate the effect of the threshold $l$ used in $\S$~\ref{sec:event_matching}, we visualize F1 scores under different threshold values in Figure \ref{fig:ablation_f1}.

\begin{table}[ht]
\centering
{\small
\begin{tabular}{lcc}
\hline
     & w/o norm. &  w/ norm. \\
\hline
w/o extract. & 4.7 & - \\
\hline
w/o PIE & 7.5 & 37.5 \\
- ARGM & 10.0 & 56.2 \\
- ARG2,3,4 & 14.6 & 73.4 \\
- ARG1 & 9.9 & 86.6 \\
\hline
\end{tabular}
}
\caption{Human evaluation for the accuracy of event matching (\%). }
\label{tab:human_eval}
\end{table}

We can observe that: 
1) Directly matching sentences to KGs (w/o extract.) has rather low performance, which necessitates the event extraction stage.
2) The event normalization step drastically improves the matching performance. 
Removing normalization step can decrease the accuracy by up to $76.7\%$.
3) In general, the matching performance gradually increases as the abstract level increases.
4) The Pearson's $r$ between automatic and human evaluation results is $0.8977$, indicating thresholding on $L2$ distance is a reasonable way to automatically filter out poorly matched events.  
Moreover, from Figure \ref{fig:ablation_f1}, we learn that event extraction, normalization, and partial information extraction improve not only performance but also robustness of event matching. 
Notably, our main model (w/ norm. -ARG1) has much higher success rate than the other models, and it is meanwhile insensitive to the tuning of threshold $l$.

\subsubsection{Effect of model structure}

We test the GNN-based reasoning model performance with different backbone text encoders (Table~\ref{tab:ablation_text_encoders}).
Compared with the baselines (``w/o know.''), our framework consistently improves performance across different versions of LMs.

We also investigate the effect of different GNN configurations in Table \ref{tab:ablation_gnn}.
Apart from the relational convolutional layers (RGCN \cite{schlichtkrull2018modeling}), we additionally test GIN \cite{xu2018powerful} and GCN \cite{kipf2016semi}, which do not model the edge type information.
We can observe that RGCN outperforms GIN and GCN under the same settings.
This indicates the discourse relation knowledge in ASER is beneficial for narrative reasoning.

We evaluate the LLM-based reasoning model under different graph sequentialization settings (Table~\ref{tab:ChatGPT_Performance}).
It is noteworthy that ChatGPT faces difficulties in understanding the knowledge represented in DOT language, resulting in a performance drop of approximately 10\%. 
One possible reason for this is that the model was not trained to comprehend such structured representations. 
Additionally, providing only node information to the model does not yield significant benefits. 
The model demonstrates improved performance when using the "Node \& Edge" representation of graphs.

\begin{table}[t]
\centering
{\small
\begin{tabular}{lccc}
\hline
\textbf{Model} & \textbf{Type} &  \textbf{w/o know.} & \textbf{w/ know.}\\
\hline
\multirow{2}{*}{BERT} &
    \multirow{1}{*}{base} & 83.63\small{$\pm$1.16} & 85.78\small{$\pm$0.74} \\
    & \multirow{1}{*}{large} & 88.85\small{$\pm$0.23} & 90.49\small{$\pm$0.41} \\
\hline
\multirow{2}{*}{RoBERTa} &
    \multirow{1}{*}{base} & 92.75\small{$\pm$0.24} & 93.30\small{$\pm$0.11} \\
    & \multirow{1}{*}{large} & 96.74\small{$\pm$0.08} & 97.10\small{$\pm$0.13} \\
\hline
\multirow{2}{*}{DeBERTa} &
    \multirow{1}{*}{base} & 96.03\small{$\pm$0.17} & 96.38\small{$\pm$0.14} \\
    & \multirow{1}{*}{large} & 98.13\small{$\pm$0.24} & 98.29\small{$\pm$0.16} \\
\hline
\end{tabular}
}
\caption{Effect of different text encoders. Three backbone language models BERT \cite{devlin2018bert}, RoBERTa \cite{liu2019roberta}, and DeBERTa \cite{he2021debertav3} are tested on SCT-v1.0.}
\label{tab:ablation_text_encoders}
\end{table}

\begin{table}[t]
\centering
{\small
\begin{tabular}{cc|cc}
\hline
 & & \multicolumn{2}{c}{$L$-layer} \\
n-hidden & conv. & 2 &  3\\
\hline
\multirow{3}{*}{128} &
    \multirow{1}{*}{\small{RGCN}} & 93.30\small{$\pm$0.11} & 92.97\small{$\pm$0.17} \\
    & \multirow{1}{*}{\small{GIN}} & 92.93\small{$\pm$0.37} & 92.57\small{$\pm$0.24} \\
    & \multirow{1}{*}{\small{GCN}} & 92.95\small{$\pm$0.10} & 93.16\small{$\pm$0.22} \\
\hline
\multirow{3}{*}{256} &
    \multirow{1}{*}{\small{RGCN}} & 93.14\small{$\pm$0.20} & 93.12\small{$\pm$0.17} \\
    & \multirow{1}{*}{\small{GIN}} & 93.05\small{$\pm$0.42} & 92.41\small{$\pm$0.31} \\
    & \multirow{1}{*}{\small{GCN}} &  92.94\small{$\pm$0.13} & 92.86\small{$\pm$0.21} \\
\hline
\end{tabular}
}
\caption{Effect of different GNN settings on SCT-v1.0.  }
\label{tab:ablation_gnn}
\end{table}

\subsection{Case study}
\label{sec:case}
\begin{figure}[!h]
\centering
\includegraphics[width=0.5\textwidth]{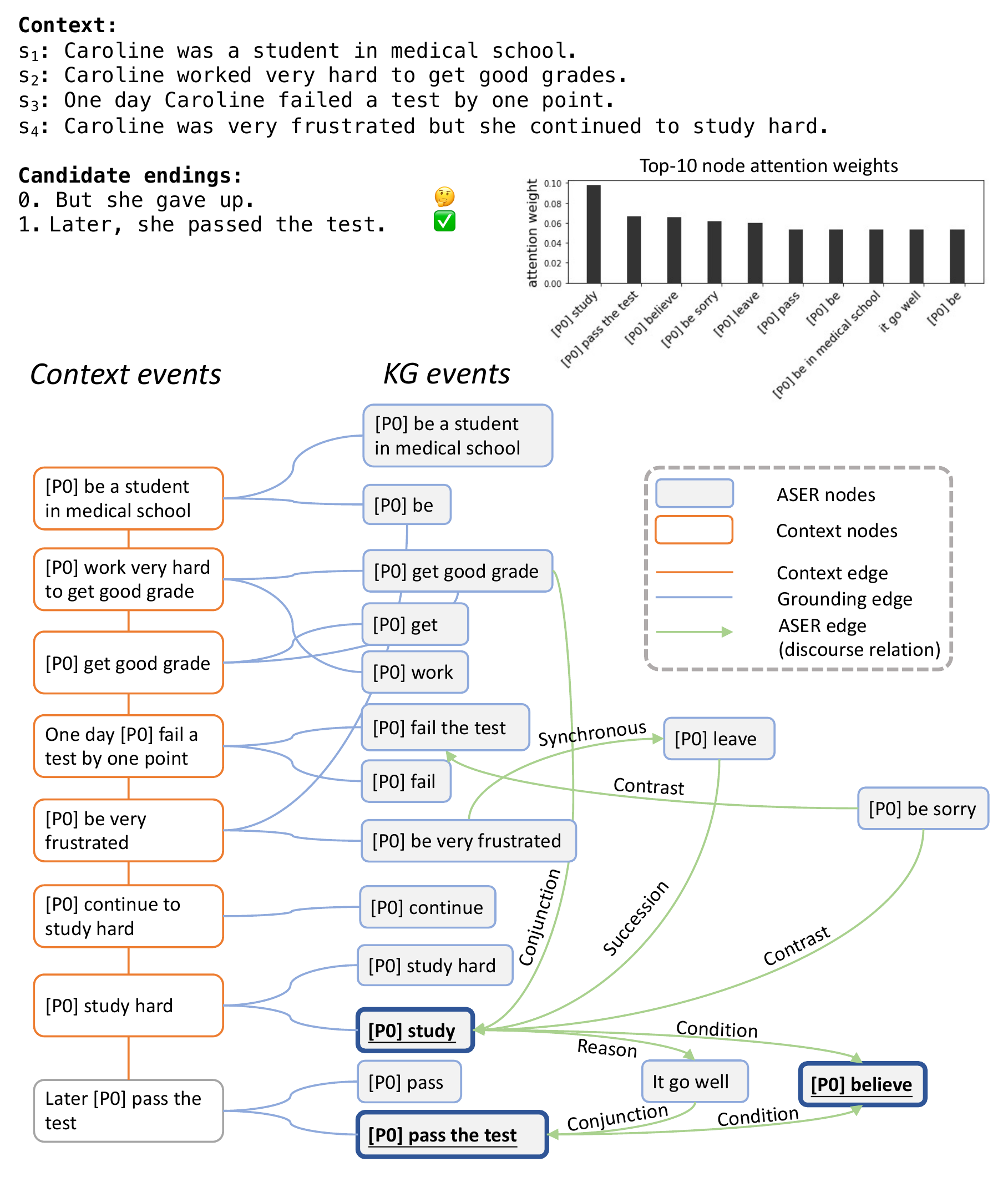}
\caption{An example from SCT-v1.0. The top-10 node attention weights are shown in the barplot. The top-3 nodes are \underline{\textbf{bolded and underlined}} .}
\label{fig:case_study}
\end{figure}

A running example is presented in Figure \ref{fig:case_study}.
The top three nodes that our model focuses on are ``\texttt{[P0]} study,'' ``\texttt{[P0]} pass the test,'' and ``\texttt{[P0]} believe.''
They are highly related to the correct candidate ending 1.
Also note the path (``\texttt{[P0]} study,'' \textit{Reason}, ``it go well,'' \textit{Conjunction},  ``\texttt{[P0]} pass the test'') could be explained as the causal story:
Someone studies hard, so it (the learning, or the exam) goes well, and he/she passes the test.

%% file: sections/limitation.tex
\section*{Limitations}
In event normalization, we only normalize personal words in event as it is the most common spans that worth normalization, normalization of other type of information are not considered, which we leave for future work.
When grounding to event-centric KGs, we consider finding the shortest paths to retrieve the knowledge subgraph due to high computational complexity of solving the Steiner tree problem.
Other retrieval methods (e.g. reinforcement learning based) could also be considered.

%% file: sections/appendix.tex
\appendix

\begin{figure*}[t]
\centering
\vspace{-10pt}
\includegraphics[width=1\textwidth]{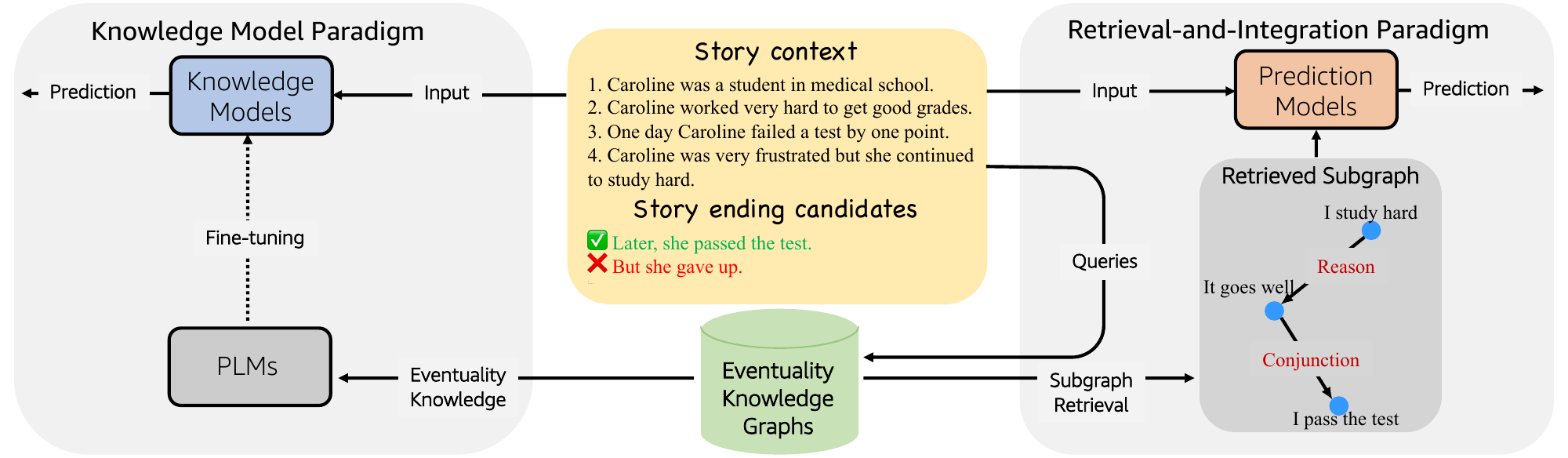}
\vspace{-10pt}
\caption{Overview of the knowledge model paradigm (left) and the retrieval-and-integration paradigm (right). The knowledge model paradigm pretrains LMs with specially designed objectives, and then further finetunes them to adapt to downstream tasks for prediction. The retrieval-and-integration paradigm retrieves relevant subgraphs of the story context and then makes predictions according to the retrieved subgraphs.}
\vspace{-10pt}
\label{fig:paradigm}
\end{figure*}

\section{Detailed experimental results}
\label{sec:appendix_detail_results}
We present the detail performance comparison for SCT-v1.0 and SCT-v1.5 (in Table \ref{tab:supplementary_sct}), as well as MCNC (in Table \ref{tab:supplementary_mcnc}).
Performance of the significant baselines in the corresponding tasks is presented.

\begin{table}[h]
\centering
\small
\scalebox{0.99}{
\begin{tabular}{l|cc}
\hline
\textbf{Method} & \textbf{SCT-v1.0} & \textbf{SCT-v1.5} \\
\hline
Random & 50.00 & 50.00 \\
\cite{chaturvedi2017story} & 77.60 & - \\
\cite{mostafazadeh2016corpus} & 58.50 & - \\
\cite{srinivasan2018simple} & 76.50 & - \\
\cite{yu2020cocolm} & 97.70 & - \\
\cite{zhou2022eventbert} & - & 91.33 \\
\cite{zhou2022claret} & - & 91.18 \\
\cite{li2019story} & 91.80 & 90.30 \\
\hline
RoBERTa-base  & 92.75\small{$\pm$0.24} & 87.14\small{$\pm$0.39} \\
RoBERTa-large  & 96.74\small{$\pm$0.08} & 92.34\small{$\pm$0.06} \\
DeBERTa-large & 98.13\small{$\pm$0.34} & 94.67\small{$\pm$0.25}  \\
\hline
\methodname\tiny{-RB} &  93.30\small{$\pm$0.11} & 87.65\small{$\pm$0.13} \\
\methodname\tiny{-RL} &    97.10\small{$\pm$0.13} & 92.86\small{$\pm$0.05} \\
\methodname\tiny{-DL} &    98.29\small{$\pm$0.16} & 95.01\small{$\pm$0.32} \\
\hline
\end{tabular}
}
\caption{Results on SCT v1.0 and v1.5. Numbers are the mean and standard deviation of accuracy (\%) over three runs.}
\label{tab:supplementary_sct}
\end{table}

\begin{table}[h]
\centering
\scalebox{0.93}{
\begin{tabular}{l|c}
\hline
\textbf{Method} & \textbf{MCNC}  \\
\hline
Random & 20.00 \\
\cite{chambers2008unsupervised} & 30.52 \\
\cite{granroth2016happens} & 49.57 \\
\cite{li2018constructing} & 52.45 \\
\cite{ding2019event} & 56.03 \\
\cite{lv2020integrating} & 58.66 \\
\cite{zhou2021modeling} & 63.62 \\
\cite{zhou2022eventbert} &  63.50\\
\cite{lee2020weakly} &           63.59 \\
\cite{lee2019multi} &           63.67 \\
\cite{zhou2022claret} & 64.61 \\
\hline
RoBERTa-base  & 61.28\small{$\pm$0.14} \\
RoBERTa-large  &  63.01\small{$\pm$0.12}\\
DeBERTa-large & 65.67\small{$\pm$0.13} \\
\hline
\methodname\tiny{-RB} &  62.11\small{$\pm$0.07} \\
\methodname\tiny{-RL} &   63.96\small{$\pm$0.15}  \\
\methodname\tiny{-DL} &    66.05\small{$\pm$0.12} \\
\hline
\end{tabular}
}
\caption{Results on MCNC. Numbers are the mean and standard deviation of accuracy (\%) over three runs.}
\label{tab:supplementary_mcnc}
\end{table}

\section{Results and statistics of event extraction and grounding}
Table \ref{tab:statistics_event_grounding} shows the detailed statistics of the event grounding and subgraph retrieval stage.
It is clear that our proposed event extraction, normalization and multi-level extraction method help alleviate the event sparsity to a large extent.
This not only reflects on the hit rate and mean L-2 distance during event grounding stage, but also in their retrieved graphs statistics.

Table \ref{tab:statistics_event_matching} shows the performance comparison between semantic similarity based matching (which we used) and the token-level similarity matching.
It is clear from the table that the token-level based similarity matching, such as tf-idf, fails to perform as good as the semantic based matching.
\label{sec:appendix_grounding}

Note that, the information extraction here is fundamentally different from the entity-centric line of work \cite{cui2021refining, cui2021incorporating, chen2022learning}, as our setting involves decomposition and semantic similarity computations over text snippets.
\begin{table}[h]
\centering
\scalebox{0.85}{
\begin{tabular}{l|cc}
\hline
& RoBERTa & BERT \\
\hline
Baseline (w/o know.) & 92.75\small{$\pm$0.24} & 83.63\small{$\pm$1.16} \\
Token-level similarity (tf-idf) & 92.84\small{$\pm$0.27} & 84.27\small{$\pm$0.73} \\
Semantic similarity (SBERT) & 93.30\small{$\pm$0.11} & 85.78\small{$\pm$0.74} \\
\hline
\end{tabular}
}
\caption{Performance comparison between baseline, token-level similarity based event matching, and semantic similarity based event matching. }
\label{tab:statistics_event_matching}
\end{table}

\begin{table*}[h]
\centering
\begin{tabular}{l|cc|cccc}
\hline
& \multicolumn{2}{c|}{\textbf{Event grounding}} & \multicolumn{4}{c}{\textbf{Subgraph retrieval}} \\
& hit rate (\%) & mean L2 distance $\bar d$ & \small $\overline{|\mathcal{V}_{sub}|}$ & \small $\overline{|\mathcal{E}_{sub}|}$ & \small $\overline{|\mathcal{V}_{joint}|}$ & \small $\overline{|\mathcal{E}_{joint}|}$ \\
\hline
w/o extract. & 1.43 & 0.9566 & 0.1235 & 0.1951 & 5.12 & 8.35  \\
\hline
\multirow{2}{*}{w/o PIE} & 88.28 & 0.3853 & 13.37 & 36.33 & 21.60 & 67.17 \\
                    & {\color{gray}12.50} & {\color{gray}0.8351} & & & & \\
\multirow{2}{*}{- ARGM} & 93.22 & 0.2819 & 22.34 & 74.12 & 30.53 & 109.64  \\
                    & {\color{gray}21.43} & {\color{gray}0.7801} & & & & \\
\multirow{2}{*}{- ARG2,3,4} & 94.38 & 0.1818 & 28.03 & 93.94 & 36.20 & 134.09  \\
                    & {\color{gray}45.44} & {\color{gray}0.6477} & & & & \\
\multirow{2}{*}{- ARG1} & 97.12 & 0.1150 & 63.27 & 281.32 & 71.41 & 330.73   \\
                    & {\color{gray}41.97} & {\color{gray}0.6968} & & & & \\
\hline
\end{tabular}
\caption{Results and statistics of event grounding and subgraph retrieval. The {\color{gray} gray numbers} are the statistics for ``w/o norm.'' experiments. }
\label{tab:statistics_event_grounding}
\end{table*}

\begin{figure}[ht]
\centering
\includegraphics[width=0.5\textwidth]{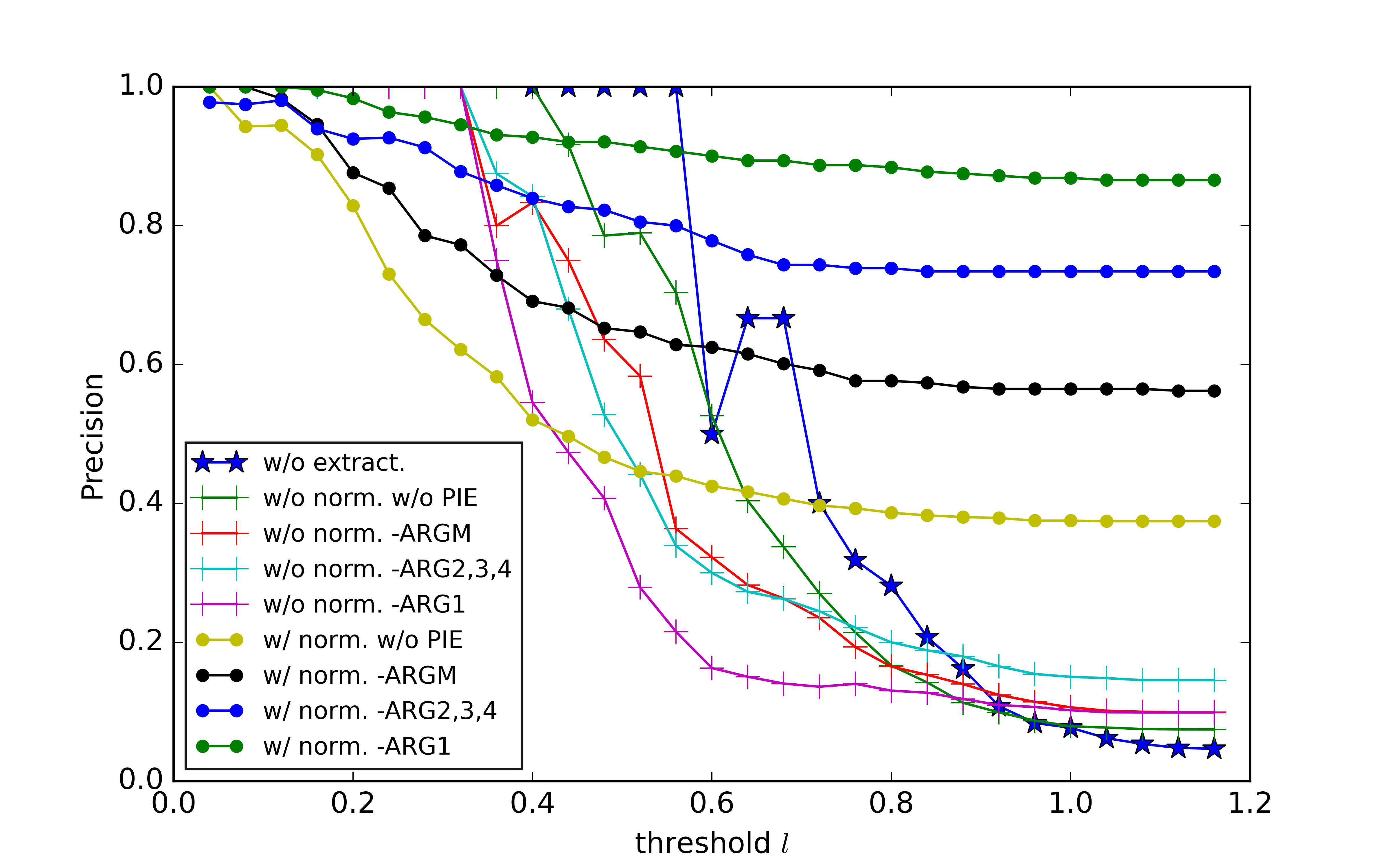}
\caption{The Precision to threshold curves. }
\label{fig:ablation_precision}
\end{figure}

\begin{figure}[ht]
\centering
\includegraphics[width=0.5\textwidth]{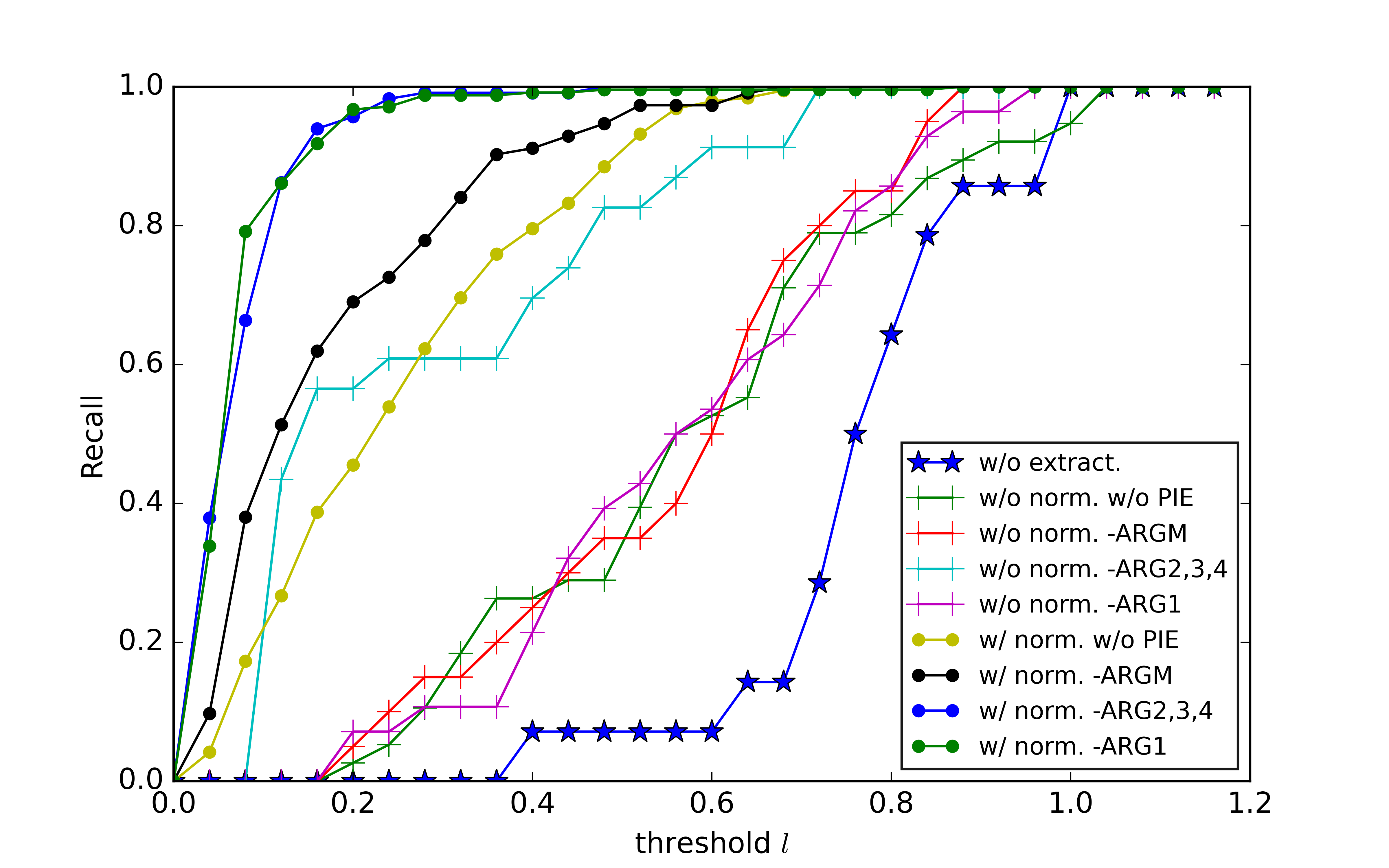}
\caption{The Recall to threshold curves. }
\label{fig:ablation_recall}
\end{figure}

\begin{figure}[ht]
\centering
\includegraphics[width=0.5\textwidth]{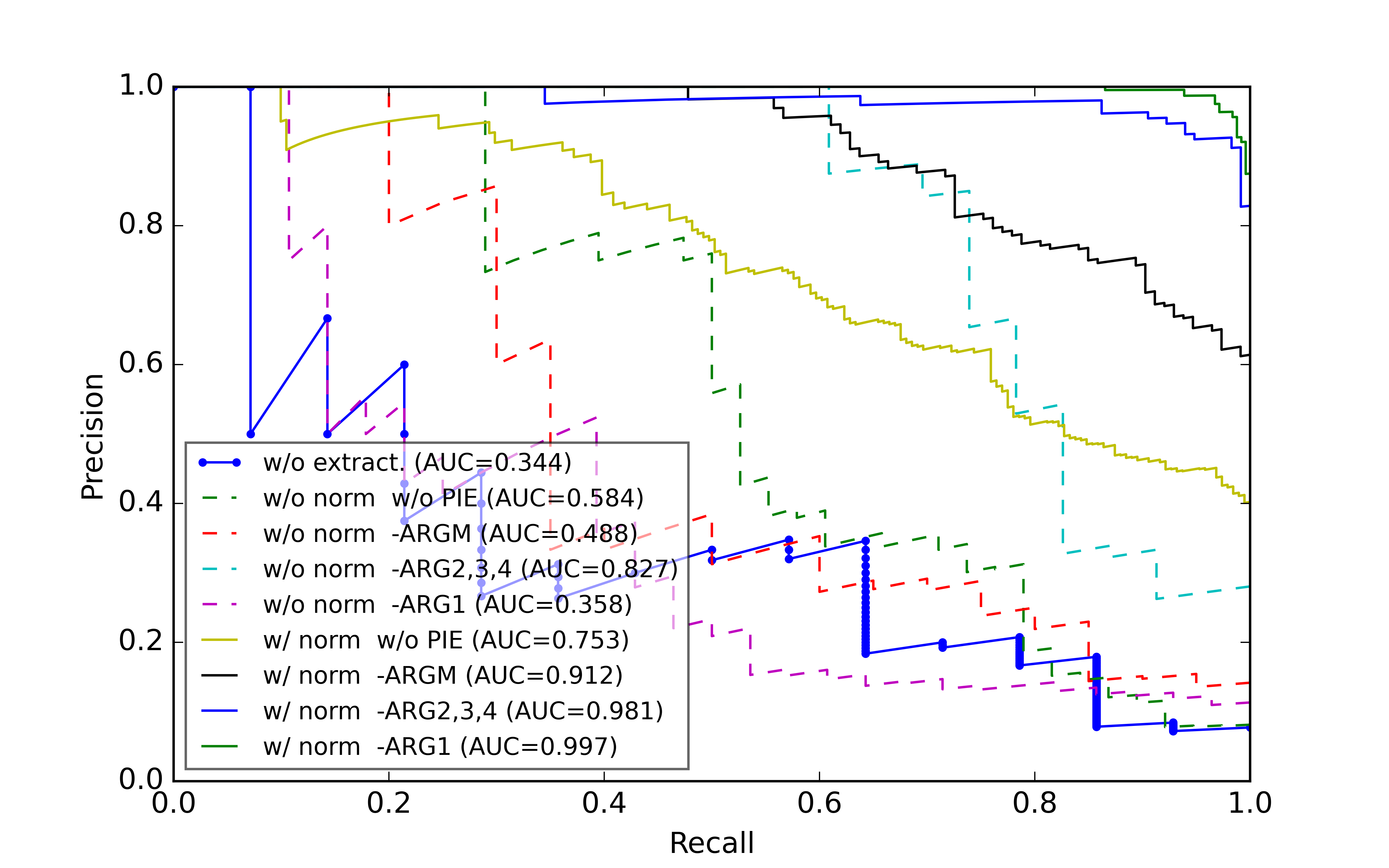}
\caption{The Precision-Recall curve.}
\label{fig:ablation_pr}
\end{figure}

\section{Supplementary case studies}
\begin{figure*}[ht]
\centering
\includegraphics[width=\textwidth]{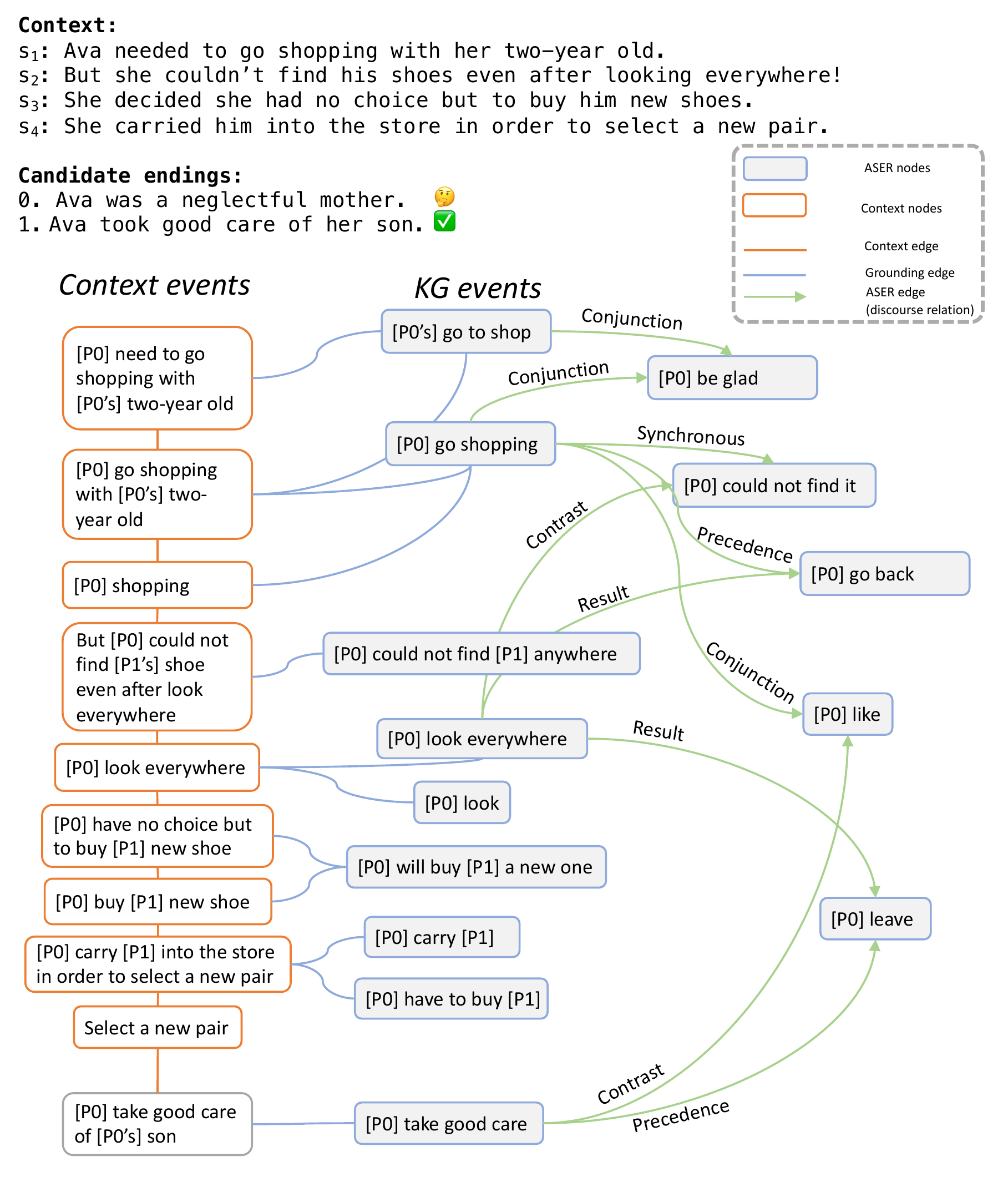}
\caption{Supplementary case 1.}
\label{fig:appendix_case_1}
\end{figure*}

\begin{figure*}[ht]
\centering
\includegraphics[width=\textwidth]{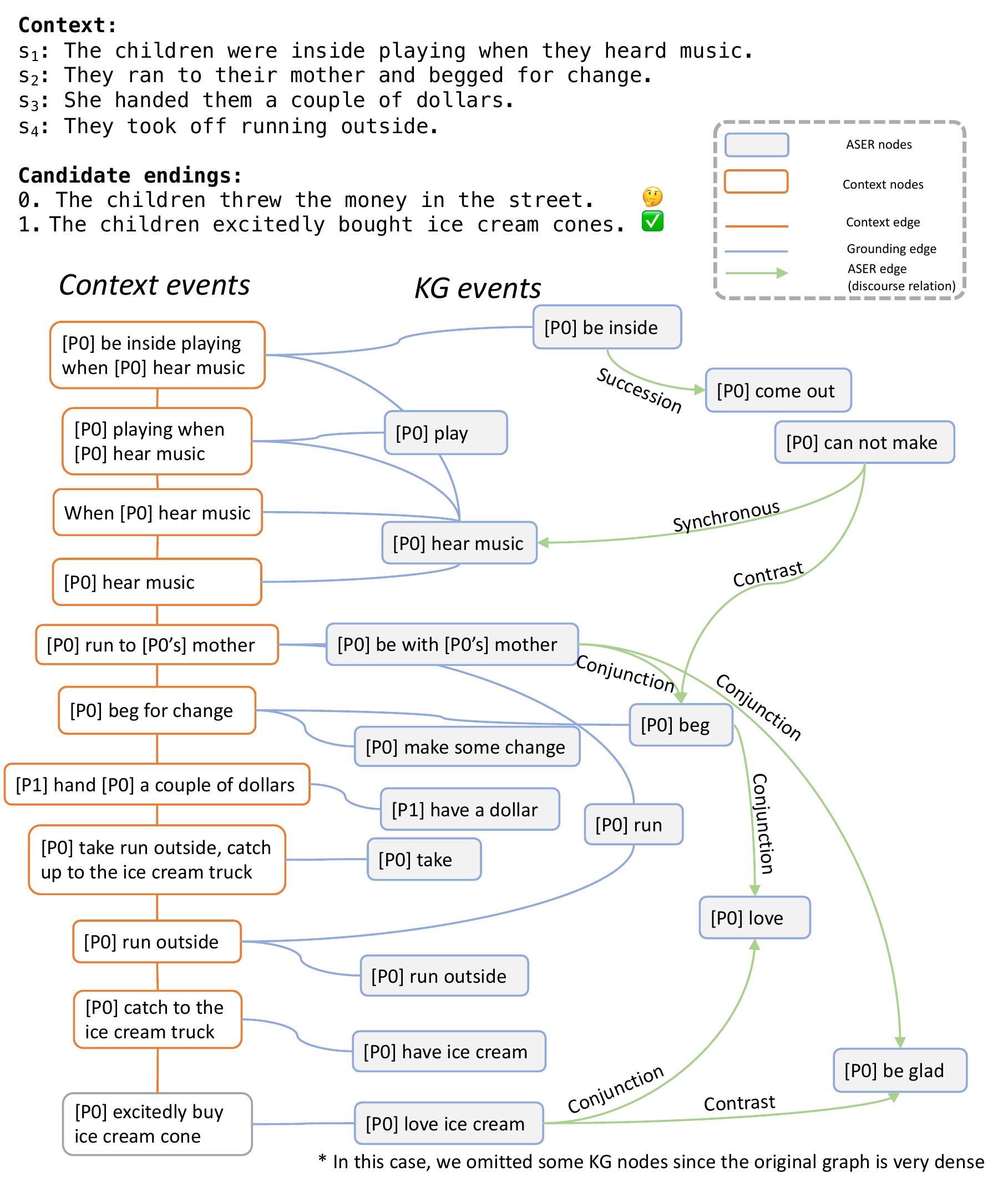}
\caption{Supplementary case 2.}
\label{fig:appendix_case_2}
\end{figure*}

Apart from the case study provided in Section \ref{sec:case}, we additionally provide another two examples in Figure \ref{fig:appendix_case_1} and \ref{fig:appendix_case_2}.

\label{sec:appendix_case}

\section{Annotation details}
\begin{figure}[ht]
\centering
\includegraphics[width=\textwidth]{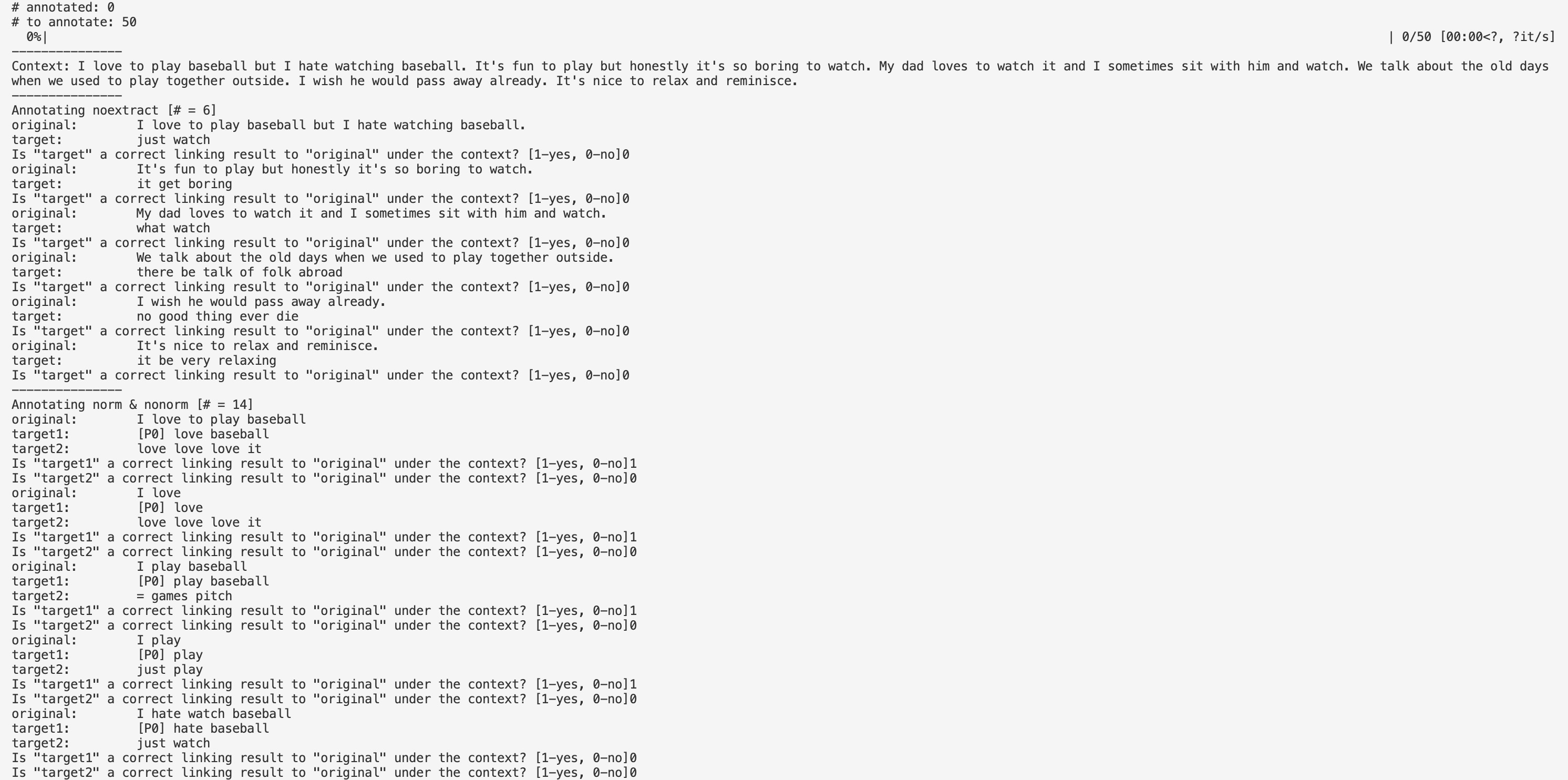}
\caption{Annotation interface in command line.}
\label{fig:ablation_annotation}
\end{figure}

We show the annotation interface presented to the expert annotators in \ref{fig:ablation_annotation}. 
Users are prompted to compare the event and its matched anchor, and then to give an evaluation of the quality (Successful-1 or Not-0).
Since the annotation requires domain-specific knowledge, we recruited 3 student researchers within our area who volunteered to help us conduct the evaluation. 
The payment to annotators is higher than the local minimum wage.

\section{Obtaining ChatGPT Performance}
\begin{table}[!t]
\small
\centering
\begin{tabular}{l c c}
\hline
\multicolumn{1}{c}{\textbf{Model}}
& \textbf{SCT-v1.0 (\%)} & \textbf{SCT-v1.5 (\%)}\\
\hline
Random                          & 50.00 & 50.00\\
ChatGPT$_\text{Prompt}$         & 77.80 & 77.00 \\
ChatGPT$_\text{w/ proscript DOT}$ & 67.80 & 69.00 \\
ChatGPT$_\text{w/ node}$         & 72.00 & \textbf{78.00} \\
ChatGPT$_\text{w/ node \& edge}$ & \textbf{79.60} & \textbf{78.00}\\
\hline
\end{tabular}
\vspace{-0.2cm}
\caption{The performance of ChatGPT performs on the SCT-v1.0 test set (sampled 500 instances) and the SCT-v1.5 validation set. The submission upload for the SCT-v1.5 leaderboard (\url{https://competitions.codalab.org/competitions/15333}) is no longer available. Therefore, we test ChatGPT performance on the validation set. The ChatGPT template is displayed in Figure~\ref{fig:ChanGpt_Template}.}
\label{tab:ChatGPT_Performance}
\vspace{-0.7cm}
\end{table}

\begin{figure*}[!t]
\centering
\includegraphics[width=\textwidth]{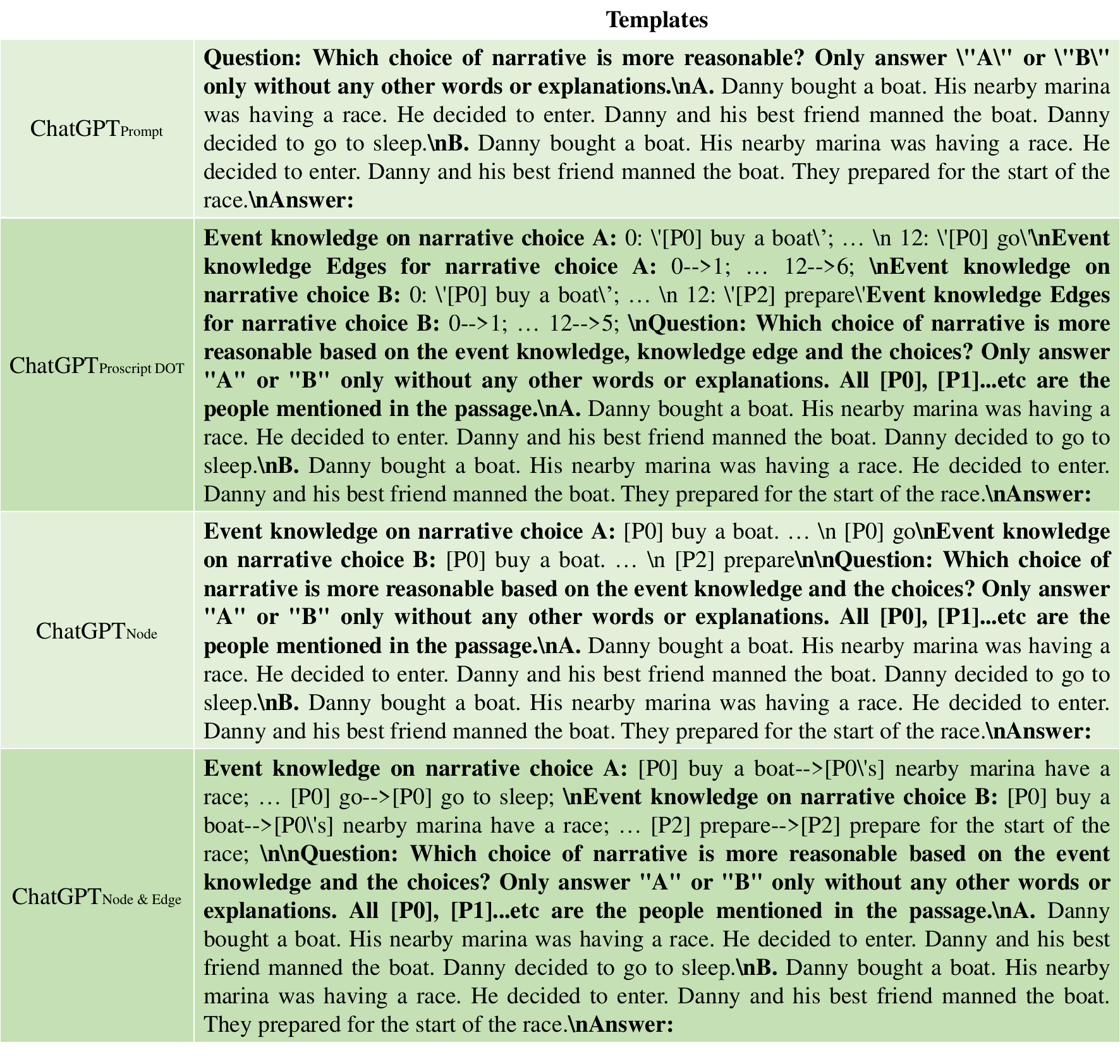}
\vspace{-0.8cm}
\caption{ChatGPT Template}
\label{fig:ChanGpt_Template}
\end{figure*}

In addition to GNNs \cite{kipf2016semi, xu2018powerful, schlichtkrull2018modeling, liu2022boosting}, we also evaluated large language models as graph reasoning modules.
Recently, large language models (e.g.,  ChatGPT~\cite{openai2022chatgpt} and GPT-4~\cite{DBLP:journals/corr/abs-2303-08774}) have shown promising performance on various tasks, and have raised concerns and discussions on topics such as factuality and privacy~\cite{wang2023survey, DBLP:journals/corr/abs-2303-12712,
DBLP:journals/corr/abs-2302-10724,
DBLP:journals/corr/abs-2304-14827,
DBLP:journals/corr/abs-2305-12870,
DBLP:journals/corr/abs-2310-10383,
DBLP:journals/corr/abs-2311-04044}.
In this paper, we test ChatGPT~\footnote{The evaluation is performed in September 2023 by calling ChatGPT Model (\textit{gpt-3.5-turbo}) API .} in narrative reasoning tasks with additional grounded knowledge. 
The zero-shot performance of large language models, which relies on the sophisticated design of templates, has shown variance across various tasks~\cite{DBLP:conf/naacl/MaZGTLZH22,
DBLP:journals/corr/abs-2309-08303,
DBLP:conf/acl/ChanLCLSWS23,
DBLP:conf/icmlc2/ChanC23}.
To obtain replicable and representative results, we follow~\citet{DBLP:conf/iclr/RobinsonW23, cheng2021question} to formulate the task as a multiple choice question answering problem.

%% file: lrec-coling_latex.bbl
\begin{thebibliography}{91}
\expandafter\ifx\csname natexlab\endcsname\relax\def\natexlab#1{#1}\fi

\bibitem[{Bach(1986)}]{bach1986algebra}
Emmon Bach. 1986.
\newblock The algebra of events.
\newblock \emph{Linguistics and philosophy}, pages 5--16.

\bibitem[{Bosselut et~al.(2019)Bosselut, Rashkin, Sap, Malaviya, Celikyilmaz, and Choi}]{DBLP:conf/acl/BosselutRSMCC19}
Antoine Bosselut, Hannah Rashkin, Maarten Sap, Chaitanya Malaviya, Asli Celikyilmaz, and Yejin Choi. 2019.
\newblock {COMET:} commonsense transformers for automatic knowledge graph construction.
\newblock In \emph{Proceedings of the 57th Conference of the Association for Computational Linguistics, {ACL} 2019, Florence, Italy, July 28- August 2, 2019, Volume 1: Long Papers}, pages 4762--4779. Association for Computational Linguistics.

\bibitem[{Bubeck et~al.(2023)Bubeck, Chandrasekaran, Eldan, Gehrke, Horvitz, Kamar, Lee, Lee, Li, Lundberg, Nori, Palangi, Ribeiro, and Zhang}]{DBLP:journals/corr/abs-2303-12712}
S{\'{e}}bastien Bubeck, Varun Chandrasekaran, Ronen Eldan, Johannes Gehrke, Eric Horvitz, Ece Kamar, Peter Lee, Yin~Tat Lee, Yuanzhi Li, Scott~M. Lundberg, Harsha Nori, Hamid Palangi, Marco~T{\'{u}}lio Ribeiro, and Yi~Zhang. 2023.
\newblock \href {http://arxiv.org/abs/2303.12712} {Sparks of artificial general intelligence: Early experiments with {GPT-4}}.
\newblock \emph{CoRR}, abs/2303.12712.

\bibitem[{Chambers and Jurafsky(2008)}]{chambers2008unsupervised}
Nathanael Chambers and Dan Jurafsky. 2008.
\newblock Unsupervised learning of narrative event chains.
\newblock In \emph{Proceedings of ACL-08: HLT}, pages 789--797.

\bibitem[{Chan and Chan(2023)}]{DBLP:conf/icmlc2/ChanC23}
Chunkit Chan and Tsz~Ho Chan. 2023.
\newblock \href {https://doi.org/10.1145/3587716.3587743} {Discourse-aware prompt for argument impact classification}.
\newblock In \emph{Proceedings of the 15th International Conference on Machine Learning and Computing, {ICMLC} 2023, Zhuhai, China, February 17-20, 2023}, pages 165--171. {ACM}.

\bibitem[{Chan et~al.(2023{\natexlab{a}})Chan, Cheng, Wang, Jiang, Fang, Liu, and Song}]{DBLP:journals/corr/abs-2304-14827}
Chunkit Chan, Jiayang Cheng, Weiqi Wang, Yuxin Jiang, Tianqing Fang, Xin Liu, and Yangqiu Song. 2023{\natexlab{a}}.
\newblock \href {https://doi.org/10.48550/arXiv.2304.14827} {Chatgpt evaluation on sentence level relations: {A} focus on temporal, causal, and discourse relations}.
\newblock \emph{CoRR}, abs/2304.14827.

\bibitem[{Chan et~al.(2023{\natexlab{b}})Chan, Liu, Chan, Cheng, Song, Wong, and See}]{DBLP:journals/corr/abs-2309-08303}
Chunkit Chan, Xin Liu, Tsz~Ho Chan, Jiayang Cheng, Yangqiu Song, Ginny~Y. Wong, and Simon See. 2023{\natexlab{b}}.
\newblock \href {https://doi.org/10.48550/arXiv.2309.08303} {Self-consistent narrative prompts on abductive natural language inference}.
\newblock \emph{CoRR}, abs/2309.08303.

\bibitem[{Chan et~al.(2023{\natexlab{c}})Chan, Liu, Cheng, Li, Song, Wong, and See}]{DBLP:conf/acl/ChanLCLSWS23}
Chunkit Chan, Xin Liu, Jiayang Cheng, Zihan Li, Yangqiu Song, Ginny~Y. Wong, and Simon See. 2023{\natexlab{c}}.
\newblock \href {https://doi.org/10.18653/v1/2023.findings-acl.4} {Discoprompt: Path prediction prompt tuning for implicit discourse relation recognition}.
\newblock In \emph{Findings of the Association for Computational Linguistics: {ACL} 2023, Toronto, Canada, July 9-14, 2023}, pages 35--57. Association for Computational Linguistics.

\bibitem[{Chaturvedi et~al.(2017)Chaturvedi, Peng, and Roth}]{chaturvedi2017story}
Snigdha Chaturvedi, Haoruo Peng, and Dan Roth. 2017.
\newblock Story comprehension for predicting what happens next.
\newblock In \emph{Proceedings of the 2017 Conference on Empirical Methods in Natural Language Processing}, pages 1603--1614.

\bibitem[{Chen et~al.(2022)Chen, Cheng, Jiang, Liu, Zhang, Shi, and Xu}]{chen2022learning}
Yi~Chen, Jiayang Cheng, Haiyun Jiang, Lemao Liu, Haisong Zhang, Shuming Shi, and Ruifeng Xu. 2022.
\newblock Learning from sibling mentions with scalable graph inference in fine-grained entity typing.
\newblock In \emph{Proceedings of the 60th Annual Meeting of the Association for Computational Linguistics (Volume 1: Long Papers)}, pages 2076--2087.

\bibitem[{Cheng et~al.(2021)Cheng, Jiang, Yang, and Xiao}]{cheng2021question}
Jiayang Cheng, Haiyun Jiang, Deqing Yang, and Yanghua Xiao. 2021.
\newblock A question-answering based framework for relation extraction validation.
\newblock \emph{arXiv preprint arXiv:2104.02934}.

\bibitem[{Cui et~al.(2021{\natexlab{a}})Cui, Yang, Cheng, and Xiao}]{cui2021incorporating}
Li~Cui, Deqing Yang, Jiayang Cheng, and Yanghua Xiao. 2021{\natexlab{a}}.
\newblock Incorporating syntactic information into relation representations for enhanced relation extraction.
\newblock In \emph{Pacific-Asia Conference on Knowledge Discovery and Data Mining}, pages 416--428. Springer.

\bibitem[{Cui et~al.(2021{\natexlab{b}})Cui, Yang, Yu, Hu, Cheng, Yi, and Xiao}]{cui2021refining}
Li~Cui, Deqing Yang, Jiaxin Yu, Chengwei Hu, Jiayang Cheng, Jingjie Yi, and Yanghua Xiao. 2021{\natexlab{b}}.
\newblock Refining sample embeddings with relation prototypes to enhance continual relation extraction.
\newblock In \emph{Proceedings of the 59th Annual Meeting of the Association for Computational Linguistics and the 11th International Joint Conference on Natural Language Processing (Volume 1: Long Papers)}, pages 232--243.

\bibitem[{Day et~al.(1998)Day, Bamford, Renandya, Jacobs, and Yu}]{day1998extensive}
Richard~R Day, Julian Bamford, Willy~A Renandya, George~M Jacobs, and Vivienne Wai-Sze Yu. 1998.
\newblock Extensive reading in the second language classroom.
\newblock \emph{RELC Journal}, 29(2):187--191.

\bibitem[{Devlin et~al.(2018)Devlin, Chang, Lee, and Toutanova}]{devlin2018bert}
Jacob Devlin, Ming-Wei Chang, Kenton Lee, and Kristina Toutanova. 2018.
\newblock Bert: Pre-training of deep bidirectional transformers for language understanding.
\newblock \emph{arXiv preprint arXiv:1810.04805}.

\bibitem[{Ding et~al.(2019)Ding, Liao, Liu, Li, and Duan}]{ding2019event}
Xiao Ding, Kuo Liao, Ting Liu, Zhongyang Li, and Junwen Duan. 2019.
\newblock Event representation learning enhanced with external commonsense knowledge.
\newblock In \emph{Proceedings of the 2019 Conference on Empirical Methods in Natural Language Processing and the 9th International Joint Conference on Natural Language Processing (EMNLP-IJCNLP)}, pages 4894--4903.

\bibitem[{Fang et~al.(2021)Fang, Zhang, Wang, Song, and He}]{fang2021discos}
Tianqing Fang, Hongming Zhang, Weiqi Wang, Yangqiu Song, and Bin He. 2021.
\newblock Discos: Bridging the gap between discourse knowledge and commonsense knowledge.
\newblock In \emph{Proceedings of the Web Conference 2021}, pages 2648--2659.

\bibitem[{Feng et~al.(2020)Feng, Chen, Lin, Wang, Yan, and Ren}]{feng2020scalable}
Yanlin Feng, Xinyue Chen, Bill~Yuchen Lin, Peifeng Wang, Jun Yan, and Xiang Ren. 2020.
\newblock Scalable multi-hop relational reasoning for knowledge-aware question answering.
\newblock In \emph{Proceedings of the 2020 Conference on Empirical Methods in Natural Language Processing (EMNLP)}, pages 1295--1309.

\bibitem[{F{\'{e}}vry et~al.(2020)F{\'{e}}vry, Soares, FitzGerald, Choi, and Kwiatkowski}]{DBLP:conf/emnlp/FevrySFCK20}
Thibault F{\'{e}}vry, Livio~Baldini Soares, Nicholas FitzGerald, Eunsol Choi, and Tom Kwiatkowski. 2020.
\newblock Entities as experts: Sparse memory access with entity supervision.
\newblock In \emph{Proceedings of the 2020 Conference on Empirical Methods in Natural Language Processing, {EMNLP} 2020, Online, November 16-20, 2020}, pages 4937--4951. Association for Computational Linguistics.

\bibitem[{Gansner et~al.(1993)Gansner, Koutsofios, North, and Vo}]{gansner1993technique}
Emden~R Gansner, Eleftherios Koutsofios, Stephen~C North, and K-P Vo. 1993.
\newblock A technique for drawing directed graphs.
\newblock \emph{IEEE Transactions on Software Engineering}, 19(3):214--230.

\bibitem[{Garey and Johnson(1977)}]{garey1977rectilinear}
Michael~R Garey and David~S. Johnson. 1977.
\newblock The rectilinear steiner tree problem is np-complete.
\newblock \emph{SIAM Journal on Applied Mathematics}, 32(4):826--834.

\bibitem[{Granroth-Wilding and Clark(2016)}]{granroth2016happens}
Mark Granroth-Wilding and Stephen Clark. 2016.
\newblock What happens next? event prediction using a compositional neural network model.
\newblock In \emph{Proceedings of the AAAI Conference on Artificial Intelligence}, volume~30.

\bibitem[{He et~al.(2021)He, Gao, and Chen}]{he2021debertav3}
Pengcheng He, Jianfeng Gao, and Weizhu Chen. 2021.
\newblock Debertav3: Improving deberta using electra-style pre-training with gradient-disentangled embedding sharing.
\newblock \emph{arXiv preprint arXiv:2111.09543}.

\bibitem[{Hoffman et~al.(2018)Hoffman, Mueller, Klein, and Litman}]{hoffman2018metrics}
Robert~R Hoffman, Shane~T Mueller, Gary Klein, and Jordan Litman. 2018.
\newblock Metrics for explainable ai: Challenges and prospects.
\newblock \emph{arXiv preprint arXiv:1812.04608}.

\bibitem[{Jiang et~al.(2023)Jiang, Chan, Chen, and Wang}]{DBLP:journals/corr/abs-2305-12870}
Yuxin Jiang, Chunkit Chan, Mingyang Chen, and Wei Wang. 2023.
\newblock \href {https://doi.org/10.48550/arXiv.2305.12870} {Lion: Adversarial distillation of closed-source large language model}.
\newblock \emph{CoRR}, abs/2305.12870.

\bibitem[{Jiayang et~al.(2023)Jiayang, Qiu, Chan, Fang, Wang, Chan, Ru, Guo, Zhang, Song et~al.}]{jiayang2023storyanalogy}
Cheng Jiayang, Lin Qiu, Tsz Chan, Tianqing Fang, Weiqi Wang, Chunkit Chan, Dongyu Ru, Qipeng Guo, Hongming Zhang, Yangqiu Song, et~al. 2023.
\newblock Storyanalogy: Deriving story-level analogies from large language models to unlock analogical understanding.
\newblock In \emph{Proceedings of the 2023 Conference on Empirical Methods in Natural Language Processing}, pages 11518--11537.

\bibitem[{Johnson et~al.(2019)Johnson, Douze, and J{\'e}gou}]{johnson2019billion}
Jeff Johnson, Matthijs Douze, and Herv{\'e} J{\'e}gou. 2019.
\newblock Billion-scale similarity search with {GPUs}.
\newblock \emph{IEEE Transactions on Big Data}, 7(3):535--547.

\bibitem[{Jordan(1998)}]{jordan1998power}
Michael~P Jordan. 1998.
\newblock The power of negation in english: Text, context and relevance.
\newblock \emph{Journal of pragmatics}, 29(6):705--752.

\bibitem[{Joshi et~al.(2020)Joshi, Chen, Liu, Weld, Zettlemoyer, and Levy}]{DBLP:journals/tacl/JoshiCLWZL20}
Mandar Joshi, Danqi Chen, Yinhan Liu, Daniel~S. Weld, Luke Zettlemoyer, and Omer Levy. 2020.
\newblock Spanbert: Improving pre-training by representing and predicting spans.
\newblock \emph{Trans. Assoc. Comput. Linguistics}, 8:64--77.

\bibitem[{Kingma and Ba(2015)}]{kingma2015adam}
Diederik~P. Kingma and Jimmy Ba. 2015.
\newblock Adam: {A} method for stochastic optimization.
\newblock In \emph{ICLR}.

\bibitem[{Kipf and Welling(2016)}]{kipf2016semi}
Thomas~N Kipf and Max Welling. 2016.
\newblock Semi-supervised classification with graph convolutional networks.
\newblock \emph{arXiv preprint arXiv:1609.02907}.

\bibitem[{Kocon et~al.(2023)Kocon, Cichecki, Kaszyca, Kochanek, Szydlo, Baran, Bielaniewicz, Gruza, Janz, Kanclerz, Kocon, Koptyra, Mieleszczenko{-}Kowszewicz, Milkowski, Oleksy, Piasecki, Radlinski, Wojtasik, Wozniak, and Kazienko}]{DBLP:journals/corr/abs-2302-10724}
Jan Kocon, Igor Cichecki, Oliwier Kaszyca, Mateusz Kochanek, Dominika Szydlo, Joanna Baran, Julita Bielaniewicz, Marcin Gruza, Arkadiusz Janz, Kamil Kanclerz, Anna Kocon, Bartlomiej Koptyra, Wiktoria Mieleszczenko{-}Kowszewicz, Piotr Milkowski, Marcin Oleksy, Maciej Piasecki, Lukasz Radlinski, Konrad Wojtasik, Stanislaw Wozniak, and Przemyslaw Kazienko. 2023.
\newblock \href {http://arxiv.org/abs/2302.10724} {Chatgpt: Jack of all trades, master of none}.
\newblock \emph{CoRR}, abs/2302.10724.

\bibitem[{Krause et~al.(2016)Krause, Xu, Uszkoreit, and Weissenborn}]{krause2016event}
Sebastian Krause, Feiyu Xu, Hans Uszkoreit, and Dirk Weissenborn. 2016.
\newblock Event linking with sentential features from convolutional neural networks.
\newblock In \emph{Proceedings of The 20th SIGNLL Conference on Computational Natural Language Learning}, pages 239--249.

\bibitem[{Lee and Goldwasser(2019)}]{lee2019multi}
I-Ta Lee and Dan Goldwasser. 2019.
\newblock Multi-relational script learning for discourse relations.
\newblock In \emph{Proceedings of the 57th Annual Meeting of the Association for Computational Linguistics}, pages 4214--4226.

\bibitem[{Lee et~al.(2020)Lee, Pacheco, and Goldwasser}]{lee2020weakly}
I-Ta Lee, Maria~Leonor Pacheco, and Dan Goldwasser. 2020.
\newblock Weakly-supervised modeling of contextualized event embedding for discourse relations.
\newblock In \emph{Findings of the Association for Computational Linguistics: EMNLP 2020}, pages 4962--4972.

\bibitem[{Lee et~al.(2021)Lee, Pacheco, and Goldwasser}]{lee2021modeling}
I-Ta Lee, Maria~Leonor Pacheco, and Dan Goldwasser. 2021.
\newblock Modeling human mental states with an entity-based narrative graph.
\newblock \emph{arXiv preprint arXiv:2104.07079}.

\bibitem[{Li et~al.(2023{\natexlab{a}})Li, Chen, Luo, Kang, Zhang, Hu, Chan, and Song}]{DBLP:journals/corr/abs-2310-10383}
Haoran Li, Yulin Chen, Jinglong Luo, Yan Kang, Xiaojin Zhang, Qi~Hu, Chunkit Chan, and Yangqiu Song. 2023{\natexlab{a}}.
\newblock \href {https://doi.org/10.48550/ARXIV.2310.10383} {Privacy in large language models: Attacks, defenses and future directions}.
\newblock \emph{CoRR}, abs/2310.10383.

\bibitem[{Li et~al.(2023{\natexlab{b}})Li, Guo, Li, Fan, Hu, Liu, Chan, Yao, and Song}]{DBLP:journals/corr/abs-2311-04044}
Haoran Li, Dadi Guo, Donghao Li, Wei Fan, Qi~Hu, Xin Liu, Chunkit Chan, Duanyi Yao, and Yangqiu Song. 2023{\natexlab{b}}.
\newblock \href {https://doi.org/10.48550/ARXIV.2311.04044} {P-bench: {A} multi-level privacy evaluation benchmark for language models}.
\newblock \emph{CoRR}, abs/2311.04044.

\bibitem[{Li et~al.(2018)Li, Ding, and Liu}]{li2018constructing}
Zhongyang Li, Xiao Ding, and Ting Liu. 2018.
\newblock Constructing narrative event evolutionary graph for script event prediction.
\newblock \emph{arXiv preprint arXiv:1805.05081}.

\bibitem[{Li et~al.(2019)Li, Ding, and Liu}]{li2019story}
Zhongyang Li, Xiao Ding, and Ting Liu. 2019.
\newblock Story ending prediction by transferable bert.
\newblock \emph{arXiv preprint arXiv:1905.07504}.

\bibitem[{Lin et~al.(2019)Lin, Chen, Chen, and Ren}]{lin2019kagnet}
Bill~Yuchen Lin, Xinyue Chen, Jamin Chen, and Xiang Ren. 2019.
\newblock Kagnet: Knowledge-aware graph networks for commonsense reasoning.
\newblock In \emph{Proceedings of the 2019 Conference on Empirical Methods in Natural Language Processing and the 9th International Joint Conference on Natural Language Processing (EMNLP-IJCNLP)}, pages 2829--2839.

\bibitem[{Liu et~al.(2022)Liu, Cheng, Song, and Jiang}]{liu2022boosting}
Xin Liu, Jiayang Cheng, Yangqiu Song, and Xin Jiang. 2022.
\newblock Boosting graph structure learning with dummy nodes.
\newblock In \emph{International Conference on Machine Learning}, pages 13704--13716. PMLR.

\bibitem[{Liu et~al.(2019)Liu, Ott, Goyal, Du, Joshi, Chen, Levy, Lewis, Zettlemoyer, and Stoyanov}]{liu2019roberta}
Yinhan Liu, Myle Ott, Naman Goyal, Jingfei Du, Mandar Joshi, Danqi Chen, Omer Levy, Mike Lewis, Luke Zettlemoyer, and Veselin Stoyanov. 2019.
\newblock Roberta: A robustly optimized bert pretraining approach.
\newblock \emph{arXiv preprint arXiv:1907.11692}.

\bibitem[{Lv et~al.(2020)Lv, Zhu, and Hu}]{lv2020integrating}
Shangwen Lv, Fuqing Zhu, and Songlin Hu. 2020.
\newblock Integrating external event knowledge for script learning.
\newblock In \emph{Proceedings of the 28th International Conference on Computational Linguistics}, pages 306--315.

\bibitem[{Ma et~al.(2022)Ma, Zhou, Gui, Tan, Li, Zhang, and Huang}]{DBLP:conf/naacl/MaZGTLZH22}
Ruotian Ma, Xin Zhou, Tao Gui, Yiding Tan, Linyang Li, Qi~Zhang, and Xuanjing Huang. 2022.
\newblock \href {https://doi.org/10.18653/v1/2022.naacl-main.420} {Template-free prompt tuning for few-shot {NER}}.
\newblock In \emph{Proceedings of the 2022 Conference of the North American Chapter of the Association for Computational Linguistics: Human Language Technologies, {NAACL} 2022, Seattle, WA, United States, July 10-15, 2022}, pages 5721--5732. Association for Computational Linguistics.

\bibitem[{Madaan and Yang(2021)}]{madaan-yang-2021-neural}
Aman Madaan and Yiming Yang. 2021.
\newblock \href {https://doi.org/10.18653/v1/2021.naacl-main.67} {Neural language modeling for contextualized temporal graph generation}.
\newblock In \emph{Proceedings of the 2021 Conference of the North American Chapter of the Association for Computational Linguistics: Human Language Technologies}, pages 864--881, Online. Association for Computational Linguistics.

\bibitem[{Min et~al.(2013)Min, Grishman, Wan, Wang, and Gondek}]{min2013distant}
Bonan Min, Ralph Grishman, Li~Wan, Chang Wang, and David Gondek. 2013.
\newblock Distant supervision for relation extraction with an incomplete knowledge base.
\newblock In \emph{Proceedings of the 2013 Conference of the North American Chapter of the Association for Computational Linguistics: Human Language Technologies}, pages 777--782.

\bibitem[{Min et~al.(2019)Min, Chen, Zettlemoyer, and Hajishirzi}]{min2019knowledge}
Sewon Min, Danqi Chen, Luke Zettlemoyer, and Hannaneh Hajishirzi. 2019.
\newblock Knowledge guided text retrieval and reading for open domain question answering.
\newblock \emph{arXiv preprint arXiv:1911.03868}.

\bibitem[{Mori et~al.(2020)Mori, Yamane, Mukuta, and Harada}]{mori2020finding}
Yusuke Mori, Hiroaki Yamane, Yusuke Mukuta, and Tatsuya Harada. 2020.
\newblock Finding and generating a missing part for story completion.
\newblock In \emph{Proceedings of the The 4th Joint SIGHUM Workshop on Computational Linguistics for Cultural Heritage, Social Sciences, Humanities and Literature}, pages 156--166.

\bibitem[{Mostafazadeh et~al.(2016)Mostafazadeh, Chambers, He, Parikh, Batra, Vanderwende, Kohli, and Allen}]{mostafazadeh2016corpus}
Nasrin Mostafazadeh, Nathanael Chambers, Xiaodong He, Devi Parikh, Dhruv Batra, Lucy Vanderwende, Pushmeet Kohli, and James Allen. 2016.
\newblock A corpus and cloze evaluation for deeper understanding of commonsense stories.
\newblock In \emph{Proceedings of the 2016 Conference of the North American Chapter of the Association for Computational Linguistics: Human Language Technologies}, pages 839--849.

\bibitem[{Mostafazadeh et~al.(2020)Mostafazadeh, Kalyanpur, Moon, Buchanan, Berkowitz, Biran, and Chu-Carroll}]{mostafazadeh2020glucose}
Nasrin Mostafazadeh, Aditya Kalyanpur, Lori Moon, David Buchanan, Lauren Berkowitz, Or~Biran, and Jennifer Chu-Carroll. 2020.
\newblock Glucose: Generalized and contextualized story explanations.
\newblock \emph{arXiv preprint arXiv:2009.07758}.

\bibitem[{Mourelatos(1978)}]{mourelatos1978events}
Alexander~PD Mourelatos. 1978.
\newblock Events, processes, and states.
\newblock \emph{Linguistics and philosophy}, 2:415--434.

\bibitem[{Murphy(2004)}]{murphy2004big}
Gregory Murphy. 2004.
\newblock \emph{The big book of concepts}.
\newblock MIT press.

\bibitem[{Nothman et~al.(2012)Nothman, Honnibal, Hachey, and Curran}]{nothman2012event}
Joel Nothman, Matthew Honnibal, Ben Hachey, and James~R Curran. 2012.
\newblock Event linking: Grounding event reference in a news archive.
\newblock In \emph{Proceedings of the 50th Annual Meeting of the Association for Computational Linguistics (Volume 2: Short Papers)}, pages 228--232.

\bibitem[{OpenAI(2023)}]{DBLP:journals/corr/abs-2303-08774}
OpenAI. 2023.
\newblock \href {http://arxiv.org/abs/2303.08774} {{GPT-4} technical report}.
\newblock \emph{CoRR}, abs/2303.08774.

\bibitem[{OpenAI(2022)}]{openai2022chatgpt}
TB~OpenAI. 2022.
\newblock Chatgpt: Optimizing language models for dialogue.
\newblock \emph{OpenAI}.

\bibitem[{Palmer et~al.(2005)Palmer, Gildea, and Kingsbury}]{palmer2005proposition}
Martha Palmer, Daniel Gildea, and Paul Kingsbury. 2005.
\newblock The proposition bank: An annotated corpus of semantic roles.
\newblock \emph{Computational linguistics}, 31(1):71--106.

\bibitem[{Peters et~al.(2019)Peters, Neumann, IV, Schwartz, Joshi, Singh, and Smith}]{DBLP:conf/emnlp/PetersNLSJSS19}
Matthew~E. Peters, Mark Neumann, Robert L.~Logan IV, Roy Schwartz, Vidur Joshi, Sameer Singh, and Noah~A. Smith. 2019.
\newblock Knowledge enhanced contextual word representations.
\newblock In \emph{Proceedings of the 2019 Conference on Empirical Methods in Natural Language Processing and the 9th International Joint Conference on Natural Language Processing, {EMNLP-IJCNLP} 2019, Hong Kong, China, November 3-7, 2019}, pages 43--54. Association for Computational Linguistics.

\bibitem[{Prasad et~al.(2008)Prasad, Dinesh, Lee, Miltsakaki, Robaldo, Joshi, and Webber}]{prasad2008penn}
Rashmi Prasad, Nikhil Dinesh, Alan Lee, Eleni Miltsakaki, Livio Robaldo, Aravind Joshi, and Bonnie Webber. 2008.
\newblock The penn discourse treebank 2.0.
\newblock In \emph{Proceedings of the Sixth International Conference on Language Resources and Evaluation (LREC'08)}.

\bibitem[{Reimers et~al.(2019)Reimers, Gurevych, Reimers, Gurevych, Thakur, Reimers, Daxenberger, Gurevych, Reimers, Gurevych et~al.}]{reimers2019sentence}
Nils Reimers, Iryna Gurevych, Nils Reimers, Iryna Gurevych, Nandan Thakur, Nils Reimers, Johannes Daxenberger, Iryna Gurevych, Nils Reimers, Iryna Gurevych, et~al. 2019.
\newblock Sentence-bert: Sentence embeddings using siamese bert-networks.
\newblock In \emph{Proceedings of the 2019 Conference on Empirical Methods in Natural Language Processing}, pages 671--688. Association for Computational Linguistics.

\bibitem[{Robinson and Wingate(2023)}]{DBLP:conf/iclr/RobinsonW23}
Joshua Robinson and David Wingate. 2023.
\newblock \href {https://openreview.net/pdf?id=yKbprarjc5B} {Leveraging large language models for multiple choice question answering}.
\newblock In \emph{The Eleventh International Conference on Learning Representations, {ICLR} 2023, Kigali, Rwanda, May 1-5, 2023}. OpenReview.net.

\bibitem[{Sakaguchi et~al.(2021)Sakaguchi, Bhagavatula, Le~Bras, Tandon, Clark, and Choi}]{sakaguchi-etal-2021-proscript-partially}
Keisuke Sakaguchi, Chandra Bhagavatula, Ronan Le~Bras, Niket Tandon, Peter Clark, and Yejin Choi. 2021.
\newblock \href {https://doi.org/10.18653/v1/2021.findings-emnlp.184} {pro{S}cript: Partially ordered scripts generation}.
\newblock In \emph{Findings of the Association for Computational Linguistics: EMNLP 2021}, pages 2138--2149, Punta Cana, Dominican Republic. Association for Computational Linguistics.

\bibitem[{Sap et~al.(2019)Sap, Le~Bras, Allaway, Bhagavatula, Lourie, Rashkin, Roof, Smith, and Choi}]{sap2019atomic}
Maarten Sap, Ronan Le~Bras, Emily Allaway, Chandra Bhagavatula, Nicholas Lourie, Hannah Rashkin, Brendan Roof, Noah~A Smith, and Yejin Choi. 2019.
\newblock Atomic: An atlas of machine commonsense for if-then reasoning.
\newblock In \emph{Proceedings of the AAAI Conference on Artificial Intelligence}, volume~33, pages 3027--3035.

\bibitem[{Schlichtkrull et~al.(2018)Schlichtkrull, Kipf, Bloem, Berg, Titov, and Welling}]{schlichtkrull2018modeling}
Michael Schlichtkrull, Thomas~N Kipf, Peter Bloem, Rianne van~den Berg, Ivan Titov, and Max Welling. 2018.
\newblock Modeling relational data with graph convolutional networks.
\newblock In \emph{European semantic web conference}, pages 593--607. Springer.

\bibitem[{Sharma et~al.(2018)Sharma, Allen, Bakhshandeh, and Mostafazadeh}]{sharma2018tackling}
Rishi Sharma, James Allen, Omid Bakhshandeh, and Nasrin Mostafazadeh. 2018.
\newblock Tackling the story ending biases in the story cloze test.
\newblock In \emph{Proceedings of the 56th Annual Meeting of the Association for Computational Linguistics (Volume 2: Short Papers)}, pages 752--757.

\bibitem[{Speer et~al.(2017)Speer, Chin, and Havasi}]{speer2017conceptnet}
Robyn Speer, Joshua Chin, and Catherine Havasi. 2017.
\newblock Conceptnet 5.5: An open multilingual graph of general knowledge.
\newblock In \emph{Thirty-first AAAI conference on artificial intelligence}.

\bibitem[{Srinivasan et~al.(2018)Srinivasan, Arora, and Riedl}]{srinivasan2018simple}
Siddarth Srinivasan, Richa Arora, and Mark Riedl. 2018.
\newblock A simple and effective approach to the story cloze test.
\newblock In \emph{Proceedings of the 2018 Conference of the North American Chapter of the Association for Computational Linguistics: Human Language Technologies, Volume 2 (Short Papers)}, pages 92--96.

\bibitem[{Sun et~al.(2019{\natexlab{a}})Sun, Bedrax-Weiss, and Cohen}]{sun2019pullnet}
Haitian Sun, Tania Bedrax-Weiss, and William Cohen. 2019{\natexlab{a}}.
\newblock Pullnet: Open domain question answering with iterative retrieval on knowledge bases and text.
\newblock In \emph{Proceedings of the 2019 Conference on Empirical Methods in Natural Language Processing and the 9th International Joint Conference on Natural Language Processing (EMNLP-IJCNLP)}, pages 2380--2390.

\bibitem[{Sun et~al.(2018)Sun, Dhingra, Zaheer, Mazaitis, Salakhutdinov, and Cohen}]{sun2018open}
Haitian Sun, Bhuwan Dhingra, Manzil Zaheer, Kathryn Mazaitis, Ruslan Salakhutdinov, and William Cohen. 2018.
\newblock Open domain question answering using early fusion of knowledge bases and text.
\newblock In \emph{Proceedings of the 2018 Conference on Empirical Methods in Natural Language Processing}, pages 4231--4242.

\bibitem[{Sun et~al.(2021)Sun, Wang, Feng, Ding, Pang, Shang, Liu, Chen, Zhao, Lu, Liu, Wu, Gong, Liang, Shang, Sun, Liu, Ouyang, Yu, Tian, Wu, and Wang}]{DBLP:journals/corr/abs-2107-02137}
Yu~Sun, Shuohuan Wang, Shikun Feng, Siyu Ding, Chao Pang, Junyuan Shang, Jiaxiang Liu, Xuyi Chen, Yanbin Zhao, Yuxiang Lu, Weixin Liu, Zhihua Wu, Weibao Gong, Jianzhong Liang, Zhizhou Shang, Peng Sun, Wei Liu, Xuan Ouyang, Dianhai Yu, Hao Tian, Hua Wu, and Haifeng Wang. 2021.
\newblock {ERNIE} 3.0: Large-scale knowledge enhanced pre-training for language understanding and generation.
\newblock \emph{CoRR}, abs/2107.02137.

\bibitem[{Sun et~al.(2019{\natexlab{b}})Sun, Wang, Li, Feng, Chen, Zhang, Tian, Zhu, Tian, and Wu}]{DBLP:journals/corr/abs-1904-09223}
Yu~Sun, Shuohuan Wang, Yu{-}Kun Li, Shikun Feng, Xuyi Chen, Han Zhang, Xin Tian, Danxiang Zhu, Hao Tian, and Hua Wu. 2019{\natexlab{b}}.
\newblock {ERNIE:} enhanced representation through knowledge integration.
\newblock \emph{CoRR}, abs/1904.09223.

\bibitem[{Verga et~al.(2020)Verga, Sun, Soares, and Cohen}]{DBLP:journals/corr/abs-2007-00849}
Pat Verga, Haitian Sun, Livio~Baldini Soares, and William~W. Cohen. 2020.
\newblock Facts as experts: Adaptable and interpretable neural memory over symbolic knowledge.
\newblock \emph{CoRR}, abs/2007.00849.

\bibitem[{Wang et~al.(2023)Wang, Liu, Yue, Tang, Zhang, Jiayang, Yao, Gao, Hu, Qi et~al.}]{wang2023survey}
Cunxiang Wang, Xiaoze Liu, Yuanhao Yue, Xiangru Tang, Tianhang Zhang, Cheng Jiayang, Yunzhi Yao, Wenyang Gao, Xuming Hu, Zehan Qi, et~al. 2023.
\newblock Survey on factuality in large language models: Knowledge, retrieval and domain-specificity.
\newblock \emph{arXiv preprint arXiv:2310.07521}.

\bibitem[{Wang et~al.(2019)Wang, Zheng, Ye, Gan, Li, Song, Zhou, Ma, Yu, Gai, Xiao, He, Karypis, Li, and Zhang}]{wang2019dgl}
Minjie Wang, Da~Zheng, Zihao Ye, Quan Gan, Mufei Li, Xiang Song, Jinjing Zhou, Chao Ma, Lingfan Yu, Yu~Gai, Tianjun Xiao, Tong He, George Karypis, Jinyang Li, and Zheng Zhang. 2019.
\newblock Deep graph library: A graph-centric, highly-performant package for graph neural networks.
\newblock \emph{arXiv preprint arXiv:1909.01315}.

\bibitem[{West et~al.(2021)West, Bhagavatula, Hessel, Hwang, Jiang, Bras, Lu, Welleck, and Choi}]{DBLP:journals/corr/abs-2110-07178}
Peter West, Chandra Bhagavatula, Jack Hessel, Jena~D. Hwang, Liwei Jiang, Ronan~Le Bras, Ximing Lu, Sean Welleck, and Yejin Choi. 2021.
\newblock \href {http://arxiv.org/abs/2110.07178} {Symbolic knowledge distillation: from general language models to commonsense models}.
\newblock \emph{CoRR}, abs/2110.07178.

\bibitem[{Wolf et~al.(2020)Wolf, Debut, Sanh, Chaumond, Delangue, Moi, Cistac, Rault, Louf, Funtowicz, Davison, Shleifer, von Platen, Ma, Jernite, Plu, Xu, Le~Scao, Gugger, Drame, Lhoest, and Rush}]{wolf-etal-2020-transformers}
Thomas Wolf, Lysandre Debut, Victor Sanh, Julien Chaumond, Clement Delangue, Anthony Moi, Pierric Cistac, Tim Rault, Remi Louf, Morgan Funtowicz, Joe Davison, Sam Shleifer, Patrick von Platen, Clara Ma, Yacine Jernite, Julien Plu, Canwen Xu, Teven Le~Scao, Sylvain Gugger, Mariama Drame, Quentin Lhoest, and Alexander Rush. 2020.
\newblock \href {https://doi.org/10.18653/v1/2020.emnlp-demos.6} {Transformers: State-of-the-art natural language processing}.
\newblock In \emph{Proceedings of the 2020 Conference on Empirical Methods in Natural Language Processing: System Demonstrations}, pages 38--45, Online. Association for Computational Linguistics.

\bibitem[{Xiong et~al.(2024)Xiong, Payani, Kompella, and Fekri}]{xiong2024large}
Siheng Xiong, Ali Payani, Ramana Kompella, and Faramarz Fekri. 2024.
\newblock Large language models can learn temporal reasoning.
\newblock \emph{arXiv preprint arXiv:2401.06853}.

\bibitem[{Xiong et~al.(2020)Xiong, Du, Wang, and Stoyanov}]{DBLP:conf/iclr/XiongDWS20}
Wenhan Xiong, Jingfei Du, William~Yang Wang, and Veselin Stoyanov. 2020.
\newblock Pretrained encyclopedia: Weakly supervised knowledge-pretrained language model.
\newblock In \emph{8th International Conference on Learning Representations, {ICLR} 2020, Addis Ababa, Ethiopia, April 26-30, 2020}. OpenReview.net.

\bibitem[{Xiong et~al.(2019)Xiong, Yu, Chang, Guo, and Wang}]{xiong2019improving}
Wenhan Xiong, Mo~Yu, Shiyu Chang, Xiaoxiao Guo, and William~Yang Wang. 2019.
\newblock Improving question answering over incomplete kbs with knowledge-aware reader.
\newblock In \emph{Proceedings of the 57th Annual Meeting of the Association for Computational Linguistics}, pages 4258--4264.

\bibitem[{Xu et~al.(2018)Xu, Hu, Leskovec, and Jegelka}]{xu2018powerful}
Keyulu Xu, Weihua Hu, Jure Leskovec, and Stefanie Jegelka. 2018.
\newblock How powerful are graph neural networks?
\newblock \emph{arXiv preprint arXiv:1810.00826}.

\bibitem[{Yang et~al.(2023)Yang, Xiong, Payani, Shareghi, and Fekri}]{yang2023harnessing}
Yuan Yang, Siheng Xiong, Ali Payani, Ehsan Shareghi, and Faramarz Fekri. 2023.
\newblock Harnessing the power of large language models for natural language to first-order logic translation.
\newblock \emph{arXiv preprint arXiv:2305.15541}.

\bibitem[{Yasunaga et~al.(2021)Yasunaga, Ren, Bosselut, Liang, and Leskovec}]{yasunaga2021qa}
Michihiro Yasunaga, Hongyu Ren, Antoine Bosselut, Percy Liang, and Jure Leskovec. 2021.
\newblock Qa-gnn: Reasoning with language models and knowledge graphs for question answering.
\newblock In \emph{Proceedings of the 2021 Conference of the North American Chapter of the Association for Computational Linguistics: Human Language Technologies}, pages 535--546.

\bibitem[{Yu et~al.(2020)Yu, Zhang, Song, and Ng}]{yu2020cocolm}
Changlong Yu, Hongming Zhang, Yangqiu Song, and Wilfred Ng. 2020.
\newblock Cocolm: Complex commonsense enhanced language model.
\newblock \emph{arXiv preprint arXiv:2012.15643}.

\bibitem[{Yu et~al.(2021)Yu, Yin, Gupta, and Roth}]{yu2021event}
Xiaodong Yu, Wenpeng Yin, Nitish Gupta, and Dan Roth. 2021.
\newblock Event linking: Grounding event mentions to wikipedia.
\newblock \emph{arXiv preprint arXiv:2112.07888}.

\bibitem[{Zhang et~al.(2022)Zhang, Liu, Pan, Ke, Ou, Fang, and Song}]{zhang2022aser}
Hongming Zhang, Xin Liu, Haojie Pan, Haowen Ke, Jiefu Ou, Tianqing Fang, and Yangqiu Song. 2022.
\newblock Aser: Towards large-scale commonsense knowledge acquisition via higher-order selectional preference over eventualities.
\newblock \emph{Artificial Intelligence}, page 103740.

\bibitem[{Zhang et~al.(2020)Zhang, Liu, Pan, Song, and Leung}]{zhang2020aser}
Hongming Zhang, Xin Liu, Haojie Pan, Yangqiu Song, and Cane Wing-Ki Leung. 2020.
\newblock Aser: A large-scale eventuality knowledge graph.
\newblock In \emph{Proceedings of the web conference 2020}, pages 201--211.

\bibitem[{Zhang et~al.(2019)Zhang, Han, Liu, Jiang, Sun, and Liu}]{DBLP:conf/acl/ZhangHLJSL19}
Zhengyan Zhang, Xu~Han, Zhiyuan Liu, Xin Jiang, Maosong Sun, and Qun Liu. 2019.
\newblock {ERNIE:} enhanced language representation with informative entities.
\newblock In \emph{Proceedings of the 57th Conference of the Association for Computational Linguistics, {ACL} 2019, Florence, Italy, July 28- August 2, 2019, Volume 1: Long Papers}, pages 1441--1451. Association for Computational Linguistics.

\bibitem[{Zhong et~al.(2022)Zhong, Liu, Ge, Mao, Jiao, Zhang, Xu, Zhu, Zeng, and Han}]{zhong2022unsupervised}
Ming Zhong, Yang Liu, Suyu Ge, Yuning Mao, Yizhu Jiao, Xingxing Zhang, Yichong Xu, Chenguang Zhu, Michael Zeng, and Jiawei Han. 2022.
\newblock Unsupervised summarization with customized granularities.
\newblock \emph{arXiv preprint arXiv:2201.12502}.

\bibitem[{Zhou et~al.(2022{\natexlab{a}})Zhou, Geng, Shen, Long, and Jiang}]{zhou2022eventbert}
Yucheng Zhou, Xiubo Geng, Tao Shen, Guodong Long, and Daxin Jiang. 2022{\natexlab{a}}.
\newblock Eventbert: A pre-trained model for event correlation reasoning.
\newblock In \emph{Proceedings of the ACM Web Conference 2022}, pages 850--859.

\bibitem[{Zhou et~al.(2021)Zhou, Geng, Shen, Pei, Zhang, and Jiang}]{zhou2021modeling}
Yucheng Zhou, Xiubo Geng, Tao Shen, Jian Pei, Wenqiang Zhang, and Daxin Jiang. 2021.
\newblock Modeling event-pair relations in external knowledge graphs for script reasoning.
\newblock In \emph{Findings of the Association for Computational Linguistics: ACL-IJCNLP 2021}, pages 4586--4596.

\bibitem[{Zhou et~al.(2022{\natexlab{b}})Zhou, Shen, Geng, Long, and Jiang}]{zhou2022claret}
Yucheng Zhou, Tao Shen, Xiubo Geng, Guodong Long, and Daxin Jiang. 2022{\natexlab{b}}.
\newblock Claret: Pre-training a correlation-aware context-to-event transformer for event-centric generation and classification.
\newblock In \emph{Proceedings of the 60th Annual Meeting of the Association for Computational Linguistics (Volume 1: Long Papers)}, pages 2559--2575.

\end{thebibliography}
